\pdfoutput=1

\documentclass[11pt]{article}

\ifx\review\undefined
\usepackage[]{acl}
\else
\usepackage[review]{acl}
\fi

\usepackage{times}
\usepackage{latexsym}
\usepackage[T1]{fontenc}
\usepackage[utf8]{inputenc}
\usepackage{microtype}
\usepackage{xspace}
\usepackage{inconsolata}
\usepackage{graphicx}
\usepackage{booktabs, multirow}
\usepackage{soul}
\usepackage{xcolor,colortbl}
\usepackage{changepage,threeparttable}
\usepackage{todonotes}
\usepackage{amsmath}
\usepackage{makecell}
\usepackage[labelfont=bf,labelsep=endash]{caption}

\usepackage{makecell}

\usepackage{graphicx}

\usepackage{multirow}
\usepackage{siunitx}

\usepackage[table]{xcolor}
\usepackage{arydshln}
\usepackage{hhline} 
\usepackage{etoolbox}
\usepackage{pgf}
\usepackage{xargs}

\definecolor{high}{HTML}{7491C2}
\definecolor{mid}{HTML}{A2B5D6}
\definecolor{low}{HTML}{C6D2E6}

\definecolor{whigh}{HTML}{76B472}
\definecolor{wmid}{HTML}{A4CDA2}
\definecolor{wlow}{HTML}{E3F0E2}

\newcommand*{\opacity}{80}

\newcommandx{\wgrad}[4][1=0.0, 2=0.5, 3=1.0]{%
    \ifdim #4 pt > #2 pt%
        \pgfmathparse{max(min(100.0*(#4-#2)/(#3-#2),100.0),0)}%
        \xdef\PercentColor{\pgfmathresult}%
        \cellcolor{high!\PercentColor!mid!\opacity}#4%
    \else
        \pgfmathparse{max(min(100.0*(#2-#4)/(#2-#1),100.0),0)}%
        \xdef\PercentColor{\pgfmathresult}%
        \cellcolor{low!\PercentColor!mid!\opacity}#4%
    \fi
}

\newcommandx{\mgrad}[4][1=0.0, 2=0.5, 3=1.0]{%
    \ifdim #4 pt > #2 pt%
        \pgfmathparse{max(min(100.0*(#4-#2)/(#3-#2),100.0),0)}%
        \xdef\PercentColor{\pgfmathresult}%
        \cellcolor{whigh!\PercentColor!wmid!\opacity}#4%
    \else
        \pgfmathparse{max(min(100.0*(#2-#4)/(#2-#1),100.0),0)}%
        \xdef\PercentColor{\pgfmathresult}%
        \cellcolor{wlow!\PercentColor!wmid!\opacity}#4%
    \fi
}

\newcommand{\lagom}{L\textsuperscript{\hspace{-3pt}A}G\textsubscript{O}M$\boldsymbol{\cdot}$NLP}

\definecolor{abscol}{HTML}{3274A1} 
\definecolor{relcol}{HTML}{E1812C} 
\definecolor{nopcol}{HTML}{3A923A} 

\newcommand{\eg}{\emph{e.g.,}}

\newcommand{\abs}{\textcolor{abscol}{\textsc{absolute}}\xspace}
\newcommand{\rel}{\textcolor{relcol}{\textsc{relative}}\xspace}
\newcommand{\nopos}{\textcolor{nopcol}{\textsc{no-pos}}\xspace}
\newcommand{\sib}{SIB-200\xspace}
\newcommand{\soroe}{SO-ROE\xspace}
\newcommand{\mblimp}{MultiBLiMP\xspace}

\newcommand{\eq}[0]{\textsuperscript{*}}

\newcommand\nonumberfootnote[1]{%
  \begingroup
  \renewcommand\thefootnote{}\footnote{#1}%
  \addtocounter{footnote}{-1}%
  \endgroup
}

\title{On the Interplay between Positional Encodings, \\Morphological Complexity, and Word Order Flexibility}

 \author{
    Kushal Tatariya\eq \and
    Wessel Poelman\eq \and
    Miryam de Lhoneux\\
  \lagom, Department of Computer Science, KU Leuven\\ 
  \texttt{\{kushaljayesh.tatariya, wessel.poelman, miryam.delhoneux\}@kuleuven.be}}

\begin{document}
\maketitle
\begin{abstract}
Language model architectures are predominantly first created for English and subsequently applied to other languages.
It is an open question whether this architectural bias leads to degraded performance for languages that are structurally different from English.
We examine one specific architectural choice: positional encodings, through the lens of the \emph{trade-off hypothesis}: the supposed interplay between morphological complexity and word order flexibility.
This hypothesis posits a trade-off between the two: a more morphologically complex language can have a more flexible word order, and vice-versa.
Positional encodings are a direct target to investigate the implications of this hypothesis in relation to language modelling.
We pretrain monolingual model variants with absolute, relative and no positional encodings for seven typologically diverse languages and evaluate them on four downstream tasks.
Contrary to previous findings, we do not observe a clear interaction between position encodings and morphological complexity or word order flexibility, as measured by various proxies.
Our results show that the choice of tasks, languages, and metrics are essential for drawing stable conclusions.
\end{abstract}

\ifx\review\undefined
\nonumberfootnote{$^*$Equal contribution. Order determined by a coin toss.}
\fi

\section{Introduction}\label{sec:intro}
A longstanding  theory in linguistics posits that human languages are roughly equally "complex", but that there exists a trade-off between the complexity of syntax and morphology \cite[\eg][]{hockett1958course,sinnemaki2014global,miestamo2017linguistic,bentz2022complexity,nijsWordOrderResponsive2025}.
The \emph{trade-off hypothesis} states that languages exhibiting higher morphological complexity tend to use less complex syntax, and vice-versa.
The hypothesis does not get a lot of attention in NLP, even though it is a useful lens through which we can investigate \emph{architectural choices} of language models for different languages.
In NLP, models are typically developed for English first, and then applied to other languages without significant architectural changes \cite{sogaard2022should}.
Subword-based transformers \cite{vaswani2017attention} are the de-facto standard architecture, with the implicit assumption that it is the best architecture for language modelling, regardless of the language.
However, some architectures might work better for some languages than for others.
We look at a specific architectural choice: \emph{positional encodings}, and their interplay with morphological complexity and word order flexibility.

\begin{figure}
        \centering
        \includegraphics[width=\linewidth]{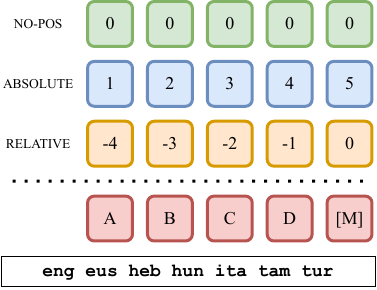}
        \caption{We investigate the relationship between positional encodings and morphology by training three model variants per language: \nopos, \abs, and \rel. We use seven typologically diverse languages, varying in morphological complexity and word order flexibility.}
        \label{fig:overview}
\end{figure}

Positional encodings in transformers signal to the model where a given token occurs in a sequence (Figure \ref{fig:overview}).
Numerous methods to encode positional information have been proposed \cite[\eg][]{liu2020learning,he2020deberta,dufter2022position}.
However, these strategies have generally been optimized for English.
This opens the question of whether these strategies work just as well for other languages with structures distinct from English.
The trade-off hypothesis gives us two areas to look for potential answers to this question: morphological complexity and word order flexibility.
Intuitively, the more flexible the word order of a language, the less useful positional encodings might be, and vice-versa.
Similarly, a model for a morphologically complex language could depend more on positional encodings if longer words are segmented into more tokens, and vice-versa.

We study the impact of three different position encoding strategies in monolingual masked language modelling across seven typologically diverse languages, specifically: no positional encodings (\nopos), relative encodings (\rel), and absolute encodings (\abs).
We aim to understand whether the importance of position encodings decreases with an increase in word order flexibility and how this relates to morphological complexity.
As a starting point we take the work of \citet{ghosh2024morphologybased}, who study the effect of \emph{removing} positional encodings from monolingual language models trained \emph{with} them.
Their results show that performance decreases from removing these encodings differ significantly between languages and that this is negatively correlated with a rough proxy of morphological complexity, namely the \emph{type-token ration (TTR)} of parallel texts.
We expand upon this line of work in three ways:
\begin{itemize}
    \item We systematically select languages with diverse typological profiles in terms of morphological complexity and word order strategies. Additionally, we make sure all languages have coverage in the downstream tasks we use and analyses we carry out. 
    \item We train monolingual models from scratch\footnote{This is important since the \emph{removal} of positional information from a trained model is notably different than the interplay between architectural choices and inherent language differences \cite[\emph{cf.}][]{ghosh2024morphologybased}.} with the three mentioned positional encoding methods, we keep all other variables constant.
    \item We analyse the impact of position encodings in the context of the trade-off hypothesis, using fine-grained proxies of morphological complexity and word order flexibility.
\end{itemize}

Our findings show that even for languages with relatively flexible word order, position encodings are essential for processing linguistic structures. Contrary to the findings of \citet{ghosh2024morphologybased}, we do not see much impact of word order flexibility or morphological complexity on the utility of position encodings. Rather, we observe the impact of position encodings to be more task specific. We also find that \emph{relative} position encodings are generally more suitable than \emph{absolute} encodings for most applications.

\section{Background}\label{sec:background}
\subsection{Position Encodings}
A transformer requires an explicit signal regarding the ordering of its input. Strategies for this include adding position encodings \cite{vaswani2017attention, devlin2019bert, takase-okazaki-2019-positional}, adding modifications to the attention matrices \cite{shaw2018selfattention, huang2020improve}, or using an RNN to process the input \cite{neishi-yoshinaga-2019-relation}. \citet{dufter2022position} identify two reference points that are commonly used to encode the position of a unit in a sentence: the \textit{absolute} position and the position of a unit \textit{relative} to other units in a sentence. Most transformers employ one or both \cite{he2020deberta, ke2021rethinking} as the basis for position encoding strategies. For both of these reference points, \citet{dufter2022position} note two methods of injecting position information into the model that are commonly associated with each. The first is "Adding Position Encodings" where position information is added to the input before it is fed to the model, and commonly uses the \textit{absolute} reference point. The second is "Modifying the Attention Matrix" which includes approaches that directly alter the attention matrix. This approach is more commonly associated with \textit{relative} position encodings. In this work, we focus on these two approaches, and compare them to a model that uses no position encodings at all. 

\subsection{Position Encodings and Linguistic~Structures}
\citet{wang2020what} show that different architectures utilise position encodings differently for various NLP tasks.  \citet{wang2020position} conduct a theoretical comparison of absolute, relative and no position encodings in BERT \cite{devlin2019bert} and show that a BERT model without position encodings is essentially a bag-of-words representation. They also show that absolute position encodings can be uninformative since any random truncations of the input can arbitrarily shift them with an offset, while relative position encodings can be beneficial for longer sequences.

Understanding the interplay of position information and the learning of linguistic structures is also done using data-level perturbations, where
\citet{sinha2021masked} randomly shuffle words and train masked language models to negate the effect of position encodings. They show shuffled language models still show a remarkable understanding of syntax and semantics, which indicates masked language models may not rely on word order information as much as we think they do. \citet{abdou2022word} extend this work to show that certain NLU tasks are sensitive to word order information. They also show results differ depending on \emph{when} data is shuffled: before or after tokenization.

While these findings are linguistically motivated, they only focus on English. 
When making architectural decisions for languages that are typologically distinct from English, these conclusions may not hold.
This is why we present an investigation into the interaction of inherent language differences and position encodings.

\subsection{Position Encodings and Multilingual~Language~Modelling}
\citet{ji2023typology} study the interplay of syntax and position encodings in a multilingual setting.
They find distances of typological feature vectors of languages to be correlated with their distance in the position space (referring to the vectors representing position information).
They leverage linguistic typology as a prior to facilitate positive transfer from high resource to low resource languages in dependency parsing. A direct investigation of the impact of various types of positional encodings on cross-lingual performance is done by \citet{ravishankar2021impact}. They find absolute position encodings work well in a multilingual language model to facilitate cross-lingual transfer. However, their experiments use synthetically perturbed English data, not "real" multilingual data.

In a monolingual setting, recent work by \citet{ghosh2024morphologybased} investigates the importance of position encodings across languages with different levels of morphological complexity.
They use monolingually pre-trained BERT-style models across 22 languages, and set positional encodings to 0 before finetuning the models on various NLI tasks. This approach does not account for the fact that the models have learned position information in the pretraining stage that may confound results. 
Using the \emph{type-token ratio (TTR)} of parallel texts as a proxy for morphological complexity, they find a negative correlation between morphological complexity and the relative drop in downstream performance between models with and without positional encodings. 
We find the use of TTR as a proxy for morphological complexity problematic, as TTR is not an inverse metric for word order flexibility.
We elaborate on this in \S\ref{sec:metrics}.

\section{Methodology}\label{sec:method}
\subsection{Language Selection and Metrics}\label{sec:metrics}
Word order flexibility can be considered as a gradient \cite[\eg][]{futrell2015quantifying,naranjo2018quantitative,levshina2023why,baylor2024multilingual}, instead of coarse groupings found in typological databases \cite{dryer2013determining}.
In an effort to create a language sample that is representative of typological features across this gradient, we sample languages based on the typological sampling framework by \citet{ploeger2025principled} using order-related features from Grambank \cite{skirgard2023grambank}.\footnote{Details are listed in \S\ref{app:sampling}.}
These coarse features allow us to create a representative language sample that is consistent across datasets.
With the languages covered by \citet{futrell2015quantifying} as a starting point, we consider languages with enough pretraining data and that are covered in potential downstream tasks as our sampling frame.\footnote{The set from which we are going to draw a sample.}
We then select a sample from our frame that covers (or saturates) the order-related typological features of interest.
This results in a sample of seven languages listed in Table \ref{tab:languages}.

\begin{table*}[ht]
    \centering
    \begin{tabular}{l|c|ccc|ccc}
    \toprule
    & & \multicolumn{3}{c|}{Word Order} & \multicolumn{3}{c}{\scalebox{0.8}{Morphological Complexity}}\\
    Language & \scalebox{0.8}{\makecell{Dominant\\ Order}} & \multicolumn{1}{c}{HDE} & \multicolumn{1}{c}{ROE} & \multicolumn{1}{c|}{\scalebox{0.8}{\soroe}} & \multicolumn{1}{c}{AV} & \multicolumn{1}{c}{$\eta$} & \multicolumn{1}{c}{\scalebox{0.8}{MATTR}} \\
    \midrule
    Basque & SOV  & \wgrad[0.07][0.28][0.50]{0.50} & \wgrad[0.03][.11][0.21]{0.21} & \wgrad[0.2][0.42][0.94]{0.72} & \mgrad[24.63][30.89][42.57]{30.22} & \mgrad[0.315][0.36][0.445]{0.38} & \mgrad[0.53][0.62][0.69]{0.68}\\
    English & SVO & \wgrad[0.07][0.28][0.50]{0.16} & \wgrad[0.03][.11][0.21]{0.03} & \wgrad[0.2][0.42][0.94]{0.20} & \mgrad[24.63][30.89][42.57]{25.20} & \mgrad[0.315][0.36][0.445]{0.34} & \mgrad[0.53][0.62][0.69]{0.53} \\
    Hebrew & SVO  & \wgrad[0.07][0.28][0.50]{0.38} & \wgrad[0.03][.11][0.21]{0.13} & \wgrad[0.2][0.42][0.94]{0.25} & \mgrad[24.63][30.89][42.57]{27.48} & \mgrad[0.315][0.36][0.445]{0.32} & \mgrad[0.53][0.62][0.69]{0.69} \\
    Hungarian & NDO  & \wgrad[0.07][0.28][0.50]{0.40} & \wgrad[0.03][.11][0.21]{0.12} & \wgrad[0.2][0.42][0.94]{0.83} & \mgrad[24.63][30.89][42.57]{42.57} & \mgrad[0.315][0.36][0.445]{0.37} & \mgrad[0.53][0.62][0.69]{0.63} \\
    Italian & SVO & \wgrad[0.07][0.28][0.50]{0.26} & \wgrad[0.03][.11][0.21]{0.06} & \wgrad[0.2][0.42][0.94]{0.27} & \mgrad[24.63][30.89][42.57]{24.63} & \mgrad[0.315][0.36][0.445]{0.33} & \mgrad[0.53][0.62][0.69]{0.58} \\
    Tamil & SOV   & \wgrad[0.07][0.28][0.50]{0.07} & \wgrad[0.03][.11][0.21]{0.09} & \wgrad[0.2][0.42][0.94]{0.94} & \mgrad[24.63][30.89][42.57]{41.50} & \mgrad[0.315][0.36][0.445]{0.45} & \mgrad[0.53][0.62][0.69]{0.69} \\
    Turkish & SOV & \wgrad[0.07][0.28][0.50]{0.22} & \wgrad[0.03][.11][0.21]{0.13} & \wgrad[0.2][0.42][0.94]{0.31} & \mgrad[24.63][30.89][42.57]{33.32} & \mgrad[0.315][0.36][0.445]{0.36} & \mgrad[0.53][0.62][0.69]{0.65} \\
    \bottomrule
\end{tabular}

    \caption{Language selection. See \S\ref{app:sampling} for details regarding the creation of this language sample. The \emph{head direction entropy (HDE)}, \emph{relation order entropy (ROE)}, and \emph{subject object ROE (\soroe)} are taken from \citet{futrell2015quantifying}. These are estimated on 1000 sentences from UD corpora. The \emph{dominant order} from \citet{dryer2013determining} is added to show there is more granularity need at this stage than coarse groupings. The measure for morphological complexity are \emph{accessor variety (AV)}, the Shannon efficiency of AV ($\eta$), and the \emph{moving average TTR (MATTR)}. These are all averages of sliding windows of 1000 tokens. All are calculated using monolingual tokenizers trained on 250k lines samples from the same dataset the models are trained on.}
    \label{tab:languages}
\end{table*}

\paragraph{Morphological Complexity.}\label{sec:morph}
For a proxy of morphological complexity, we use the \emph{Accessor Variety (AV)} and \emph{Accessor Efficiency ($\eta$)} metrics introduced by \citet{poelman2025confounding}.
AV
measures how often types from the vocabulary of a tokenizer co-occur, either on the left or right in a corpus.
$\eta$ measures the Shannon efficiency of this distribution.
We follow \citeauthor{poelman2025confounding} and use the \emph{right accessor} (successor) for both.
Intuitively, AV and $\eta$ capture some properties of morphological complexity when using a subword tokenizer: a language that uses many affixes that can occur in different places results in more \emph{choice} or \emph{ambiguity} from one token to another; a proxy for complexity.
Similar to the \emph{moving average TTR} \cite[MATTR, ][]{covington2010cutting}, we calculate AV in a sliding widow of tokens to combat the effect of text length. 

\citet{ghosh2024morphologybased} approximate morphological complexity by calculating TTR on FLORES-200 using words or characters.
This creates a mismatch between (1) the units a model sees (subword tokens) and what the proxy is calculated on (words or characters), and (2) the data that is used for model training and what data the proxy is calculated on.
We remove these confounds by maintaining consistency across units and corpora.
For the sake of completeness and comparison with \citeauthor{ghosh2024morphologybased}, we include MATTR using our tokenizers.

\paragraph{Word Order Flexibility.}

For word order flexibility, we use the UD-based metrics introduced by \citet{futrell2015quantifying}.
\emph{Head direction entropy (HDE)} measures the conditional entropy of whether a head is to the right or left of a dependent.
It is a proxy for consistency in head direction, giving us some indication of word order flexibility once conditioned on all relation types and part of speech labels.
\emph{Relation order entropy (ROE)} is the conditional entropy of the order of words in a local sub-tree.
This measure is sensitive to the size of a corpus.
To alleviate this, it can be restricted to specific relations of interest, such as the position of the subject and object in the main clause.
This is referred to as the \emph{subject object ROE (\soroe)}.

We want to emphasize a point made by \citet{paul-kiparsky} and reinforced by \citet{futrell2015quantifying}: languages with flexible word order (\soroe higher than 0.625, for instance Tamil, Basque and Hungarian, see Table \ref{tab:languages}) tend to use case marking.
Thus, a more flexible word order implies the presence of case marking.
However, inversely, the presence of case marking does not necessarily imply a flexible word order. 
Turkish, for example, has low \soroe and HDE, indicating a more rigid word order, but it does use case markings and is considered in the NLP landscape a more "morphologically complex" language.
Put differently, a metric for morphological complexity is not directly an "inverse" metric for word order flexibility (see Table \ref{tab:languages}).
Therefore, using measures like TTR or MATTR, as a rough proxy for word order flexibility (as is done in \citet{ghosh2024morphologybased}) can be problematic, and the two must not be conflated.

With HDE and \soroe, unfortunately, we also introduce the confound we alluded to earlier: the units of this measure are not the same as the units the model sees.
Still, it gives a more realistic and gradient view of word order flexibility than shuffling English \cite{ravishankar2021impact,abdou2022word}, using typological groupings \cite{dryer2013determining}, or using a proxy for morphological complexity as an inverse metric for word order flexibility \cite{ghosh2024morphologybased}.

\begin{table*}[ht]
    \centering
     \resizebox{\linewidth}{!}{    \begin{tabular}{lll}
        \toprule
        \textbf{Experimental setting} & \textbf{\citealt{ghosh2024morphologybased}} & \textbf{Ours} \\
        \midrule
        Languages studied & 6-22, depending on the task & 7 sampled for diversity and consistency across tasks \\
        Tasks & POS, UD, XNLI, NER, PAWS-X & UD, NER, \sib, \mblimp \\
        Model comparison & Removing vs keeping positional encodings & Models with three strategies trained from scratch \\ 
        Morphological complexity & TTR & AV, $\eta$, MATTR\\
        Word order flexibility & TTR & HDE, SO-ROE \\ 
        Metric unit & Words or characters & Subwords from tokenizer \\
        Corpus used for metrics & 2k lines from FLORES-200 & 250k lines from tokenizer and model training data \\ 
        \bottomrule
\end{tabular}
}
    \caption{Comparison of experimental conditions.}
    \label{tab:comparison}
\end{table*}

\subsection{Models and Tokenization}
We use the training setup from \citet{tatariya2024how}.
For each of our seven languages, we train three monolingual RoBERTa-base models \cite{liu2019roberta}, one for each of the positional encoding strategies, resulting in 21 models total.
We choose not to remove positional encodings from a model after pretraining \cite{ghosh2024morphologybased} since we are interested in inherent model architecture choices and their relation to inherent language characteristics.
We keep hyperparameters and other settings as consistent as possible.
Our tokenizers are monolingual BPE \cite{gage1994new,sennrich2016neural} tokenizers.
\footnote{Details are listed in \S\ref{app:setup}.}

\begin{figure*}[ht]
    \centering
    \includegraphics[width=0.24\linewidth]{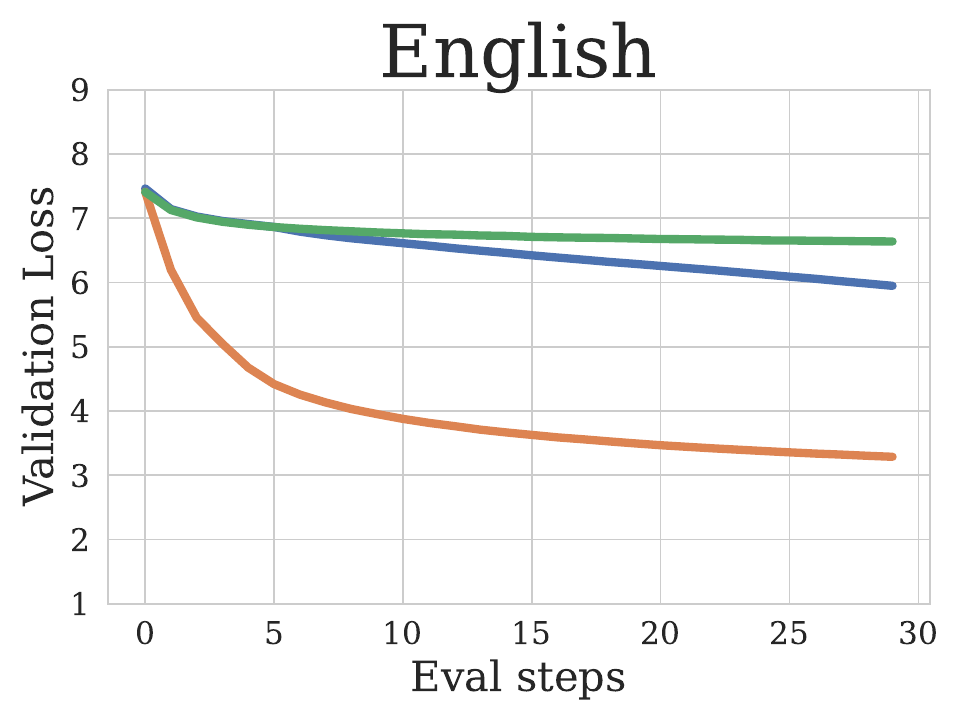}
    \includegraphics[width=0.24\linewidth]{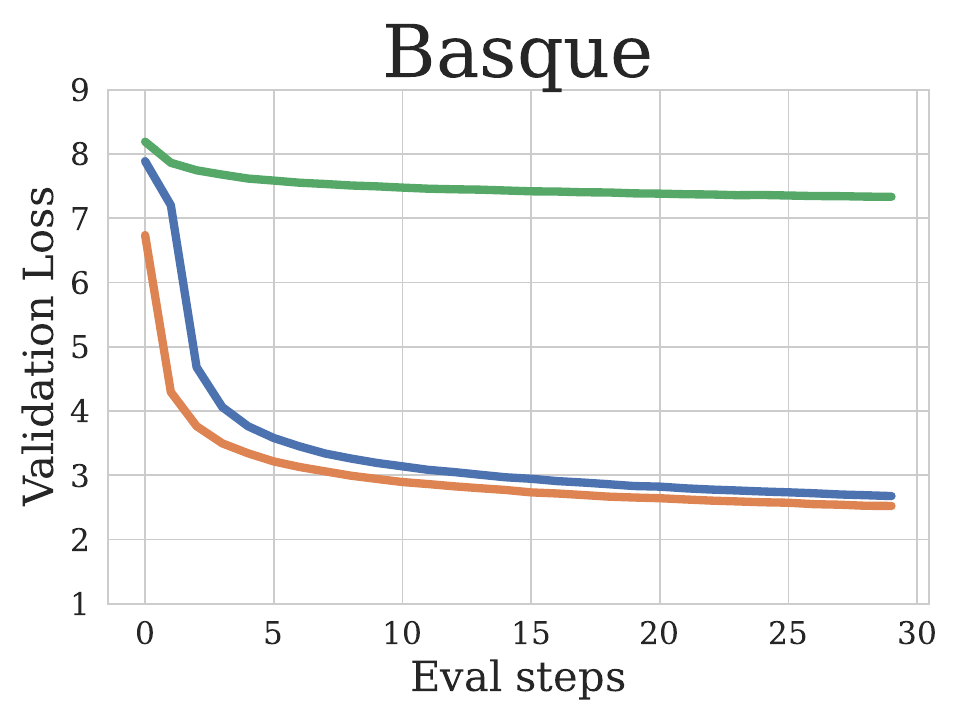}
    \includegraphics[width=0.24\linewidth]{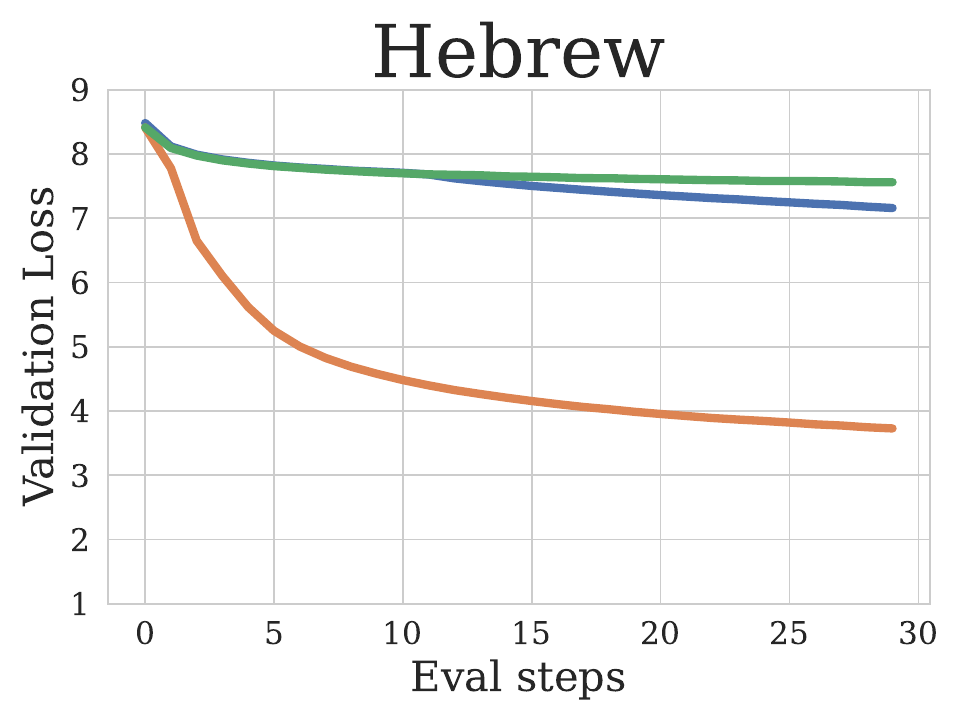}
    \includegraphics[width=0.24\linewidth]{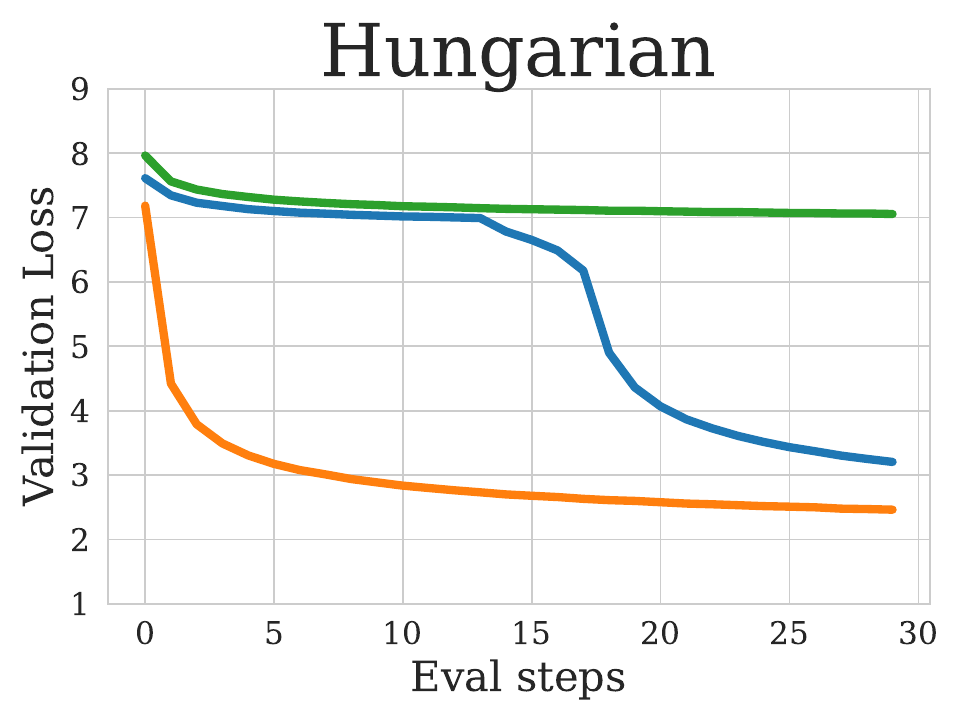}
    \includegraphics[width=0.24\linewidth]{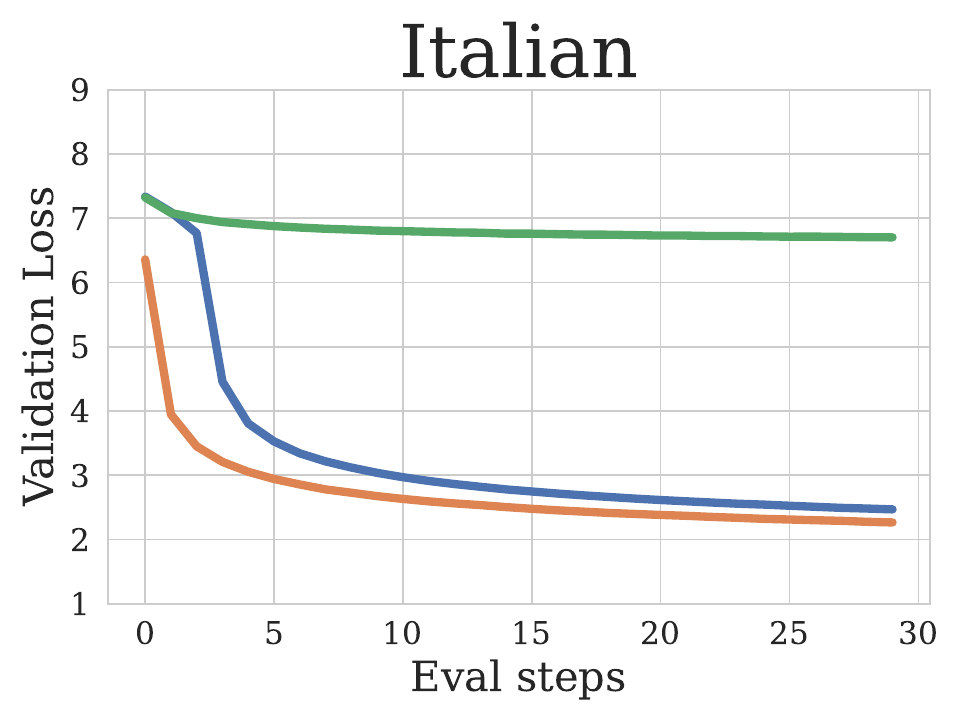}
    \includegraphics[width=0.24\linewidth]{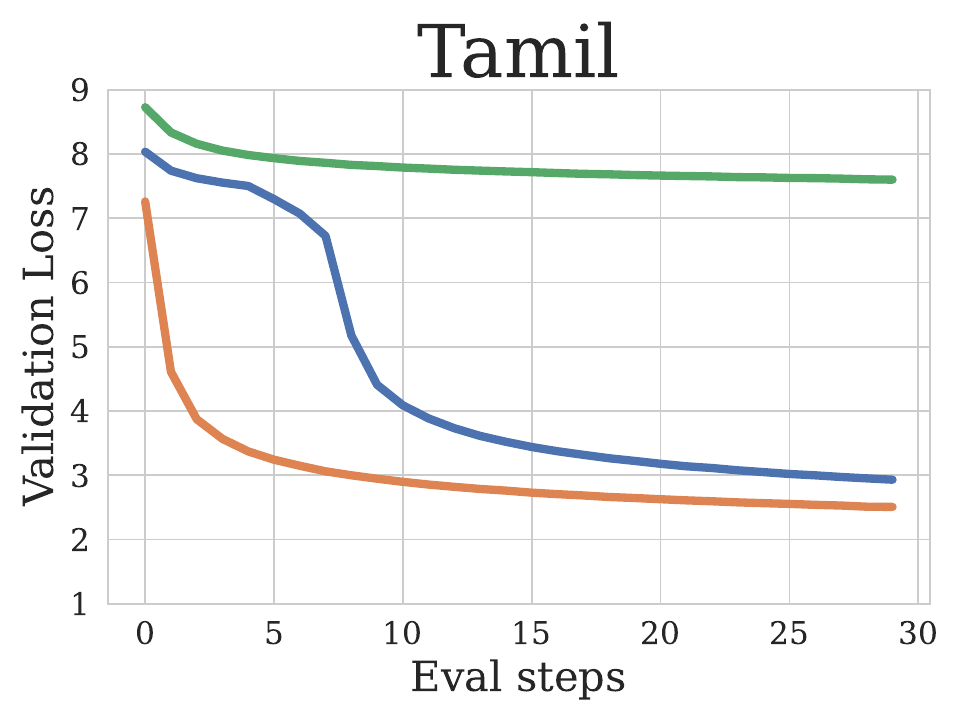}
    \includegraphics[width=0.24\linewidth]{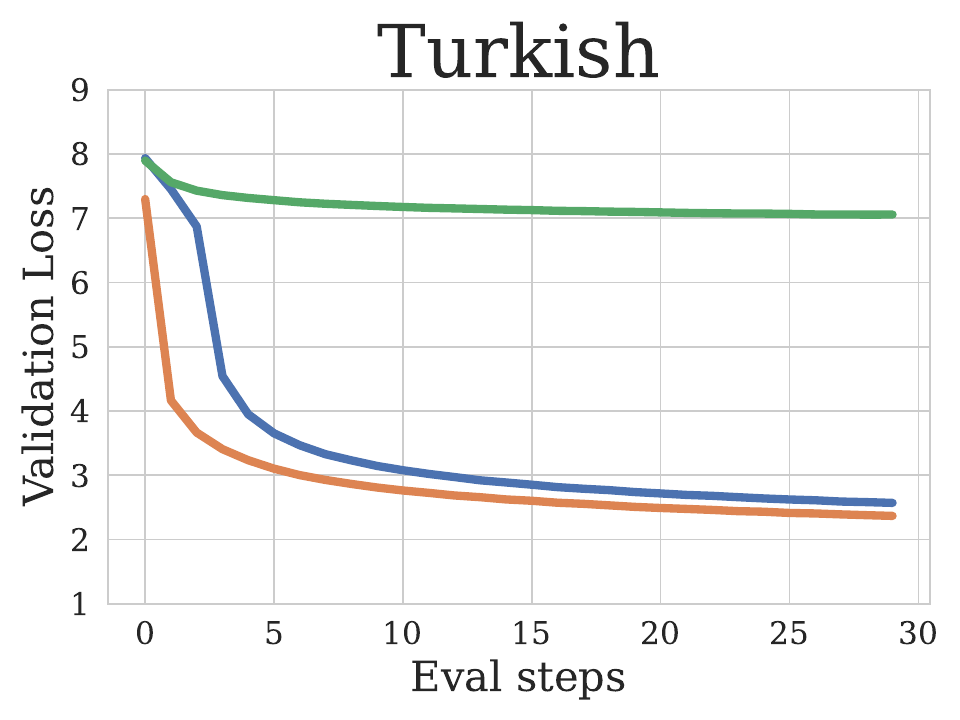}
    \caption{Loss curves on validation set for pretraining for \abs, \rel, and \nopos.}
    \label{fig:loss}
\end{figure*}

\subsection{Data and Evaluation}
For training our tokenizers and pretraining our monolingual models, we sample one hundred million lines per language from \texttt{FineWeb} \cite[English only, ][]{penedo2024fineweb} and \texttt{FineWeb-2} \cite[the other languages, ][]{penedo2025fineweb2}.
We sample the same amount of data to remove the confound of data disparity between languages.
Similarly, we choose \emph{lines} instead of \emph{tokens}, as the latter is confounded by language characteristics \cite{poelman2025confounding} and circular: we need a tokenizer to count tokens to create a sample, but we need a sample to train our tokenizers.
Our corpus-based metrics are also calculated on 250k lines randomly sampled from this dataset.

We use the following downstream tasks:
\begin{itemize}
    \item Dependency Parsing using Universal Dependencies \cite[UD v2.15,][]{nivre2020universal,demarneffe2021universal,zeman2024universal} represents a challenging token- and sentence-level task.
    \item Named-Entity Recognition (NER) using WikiAnn \cite{pan-etal-2017-cross, rahimi-etal-2019-massively}, which represents an in-between: named entities can be multi-word expressions or single tokens.
    \item \sib \cite{adelani2024sib200} is a simple, sentence-level text classification task. 
    \item \mblimp 1.0 \cite{jumelet2025multiblimp} is a minimal pair, sentence level task where ungrammatical sentences should have higher (worse) perplexity values than grammatical ones. The dataset specifically targets \mbox{\emph{agreement}}, which represents an interesting setting for both word order and morphological features. Since we use encoder-models, we use pseudo-perplexity \cite{salazar2020masked,kauf2023better}, as implemented in the \texttt{minicons} library \cite{misra2022minicons}.
\end{itemize}

We selected these tasks to represent both token-level and sentence-level tasks, as well as to cover different difficulty levels.\footnote{We experimented with Belebele \cite{bandarkar2024belebele}, a challenging reading comprehension task, but found that none of the models scored above the majority choice. 
}
In contrast with \citet{ghosh2024morphologybased}, who used a subset of XTREME \cite{hu2020xtreme} covering different languages per task (see \citet{ploeger2024what}), in our setup all languages are available for all tasks, making comparisons easier and conclusions more consistent.
Table \ref{tab:comparison} summarizes the differences between our and \citeauthor{ghosh2024morphologybased}'s study.

\section{Results}\label{sec:results}

In this section, we discuss \emph{general} results of the monolingual masked language modelling performance with \abs, \rel and \nopos, followed by their respective downstream results.
In \S\ref{sec:analysis} we go into more detail regarding individual language differences.

\begin{figure*}[ht]
    \centering
    \includegraphics[width=0.24\linewidth]{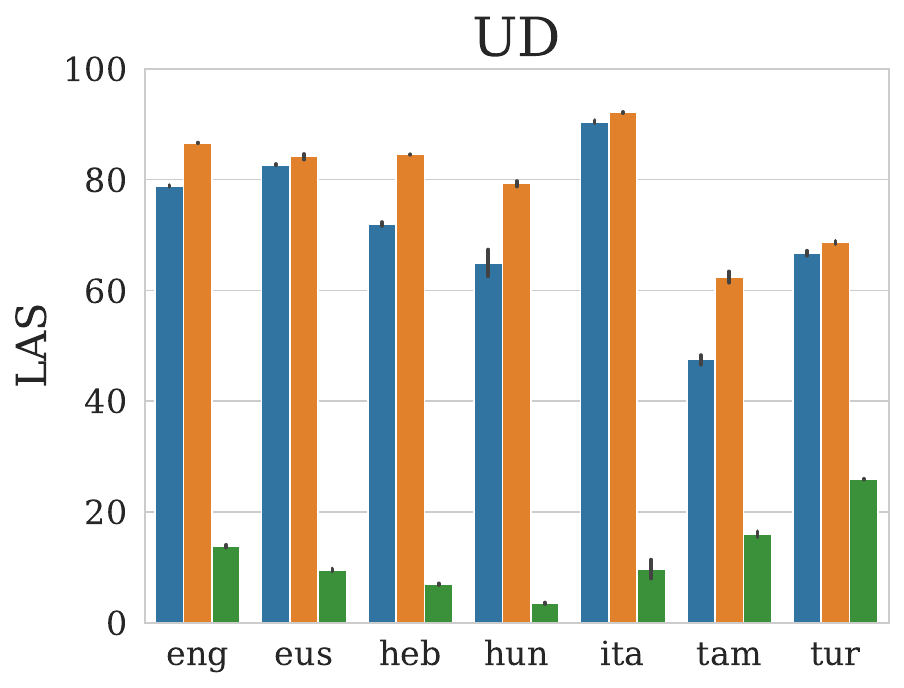}
    \includegraphics[width=0.24\linewidth]{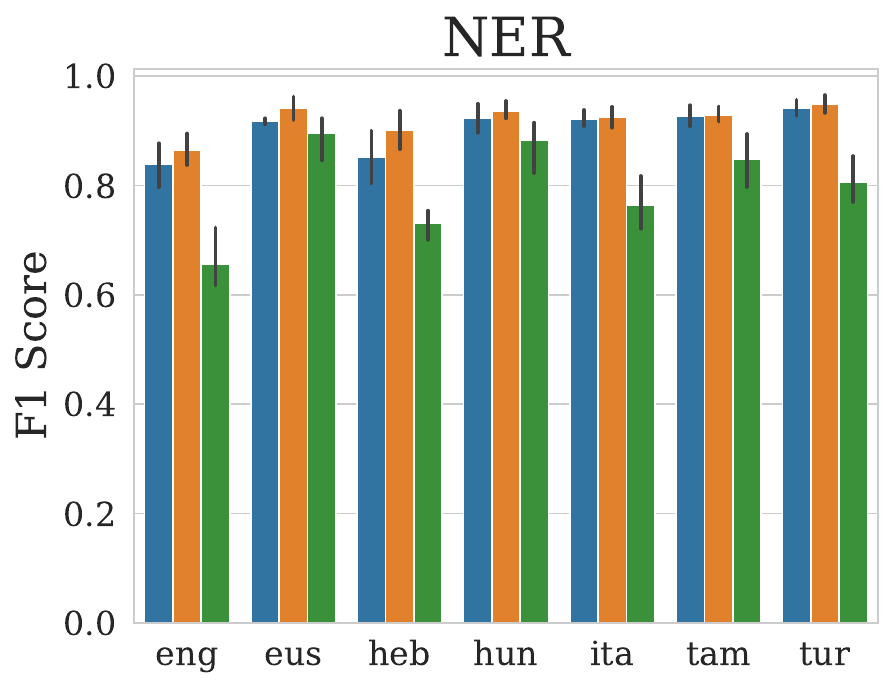}
    \includegraphics[width=0.24\linewidth]{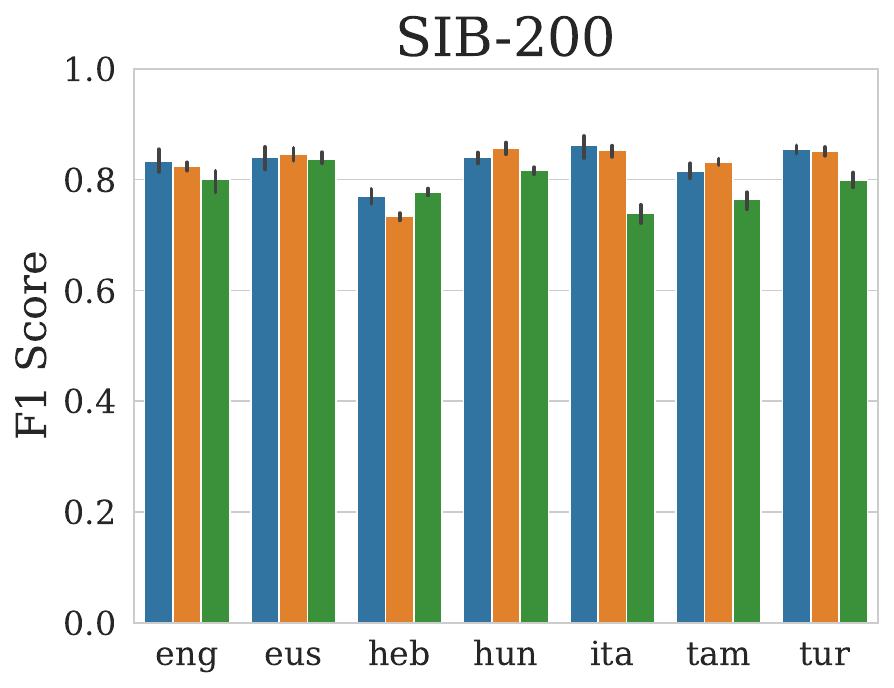}
    \includegraphics[width=0.24\linewidth]{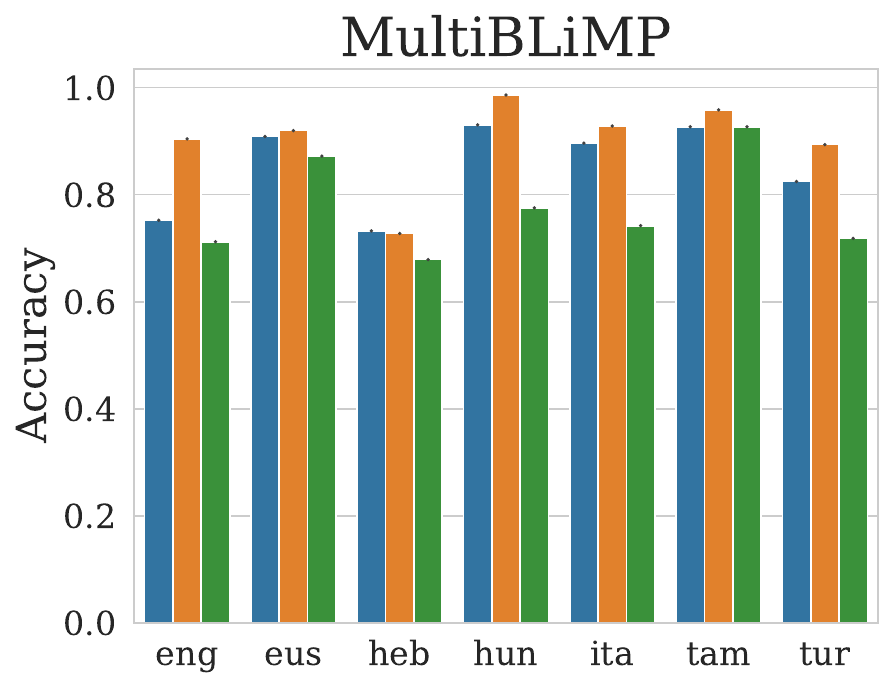}
    \includegraphics[width=0.4\linewidth]{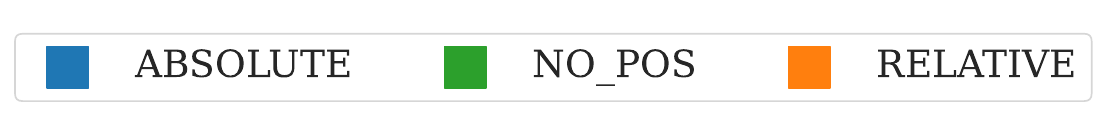}
    \caption{Results per task, language, and positional encoding type. Scores are averaged over 5 runs. Full results are in \S\ref{app:full-results}.}
    \label{fig:results}
\end{figure*}

\subsection{Pretraining}
The loss curves for the \texttt{FineWeb-2} validation set of each language for each position encoding type are shown in Figure \ref{fig:loss}. 
With the same hyperparameters, none of the \nopos models achieve a loss below 6.\footnote{It could be argued that the hyperparameters we used are tailored to the models \emph{with} positional encodings. While we try to keep the comparison as fair as possible, we did not fully explore the effect of hyperparameters. We discuss this more in the limitations (\S\ref{sec:limitations}).}
All languages also have better losses with \rel than with \abs. 
Additionally, the models achieve a lower loss more quickly with \rel compared to \abs.
Better losses with \rel encodings indicate that \citet{huang2020improve}'s findings are applicable to languages beyond English. 
Still, the learning dynamics are quite different between the languages; English and Hebrew do not converge with \abs, for example, and their losses for \rel are higher than the others.
Similarly, Hungarian and Tamil take noticeably longer to start converging with \abs than with \rel.

\subsection{Finetuning}
Results for the finetuned models on downstream tasks are in Figure \ref{fig:results}.
We see that the importance of position encodings differs more per task than language. This echoes the findings from \citet{wang2020what} and \citet{wang2020position}.
The difference between \rel and \abs for most tasks is, surprisingly, less pronounced than the difference of the loss in Figure \ref{fig:loss} (e.g., for English and Hebrew).
This means the models are able to "catch up" during finetuning, even if they did not fully converge during pretraining.

Dependency parsing requires syntactic knowledge and therefore needs access to positional information. We can see that the \nopos models are drastically worse than \abs and \rel.
Thus, position encodings are necessary for the model to learn syntax, regardless of the morphological complexity or word order flexibility of the language.
Particularly, \rel fares the best among the three strategies, reflective of the lower losses seen in Figure \ref{fig:loss}.

We see similar trends for NER, where some languages have noticeably lower performance with \nopos than others, while \rel position encodings generally have the strongest performance.
However, NER does not require a lot of syntactic or morphological understanding, apart from identifying the multi-word named entities, or those that may be inflected in some languages.

With \sib we observe a different pattern. The gap between the three variants is generally not as pronounced as with the other tasks.
\sib is a classification task that does not require much understanding of syntax, making position encodings not essential.

On \mblimp, languages like Basque and Tamil show roughly the same performance across the three model variants.
For English, Hungarian, and Turkish, \rel strongly outperforms \abs and \nopos.
The \mblimp results somewhat mirror the UD results, albeit with much higher scores for the \nopos model variant (it is ultimately a sentence-level task).
Since both datasets are highly dependent on both word order and morphology, they should be ideal candidates to show the results found by \citeauthor{ghosh2024morphologybased}
We will discuss this in the next section.

\section{Analysis}\label{sec:analysis}

\subsection{Position Encodings and Morphological~Complexity}
\begin{figure}[ht]
    \centering
    \includegraphics[width=0.45\linewidth]{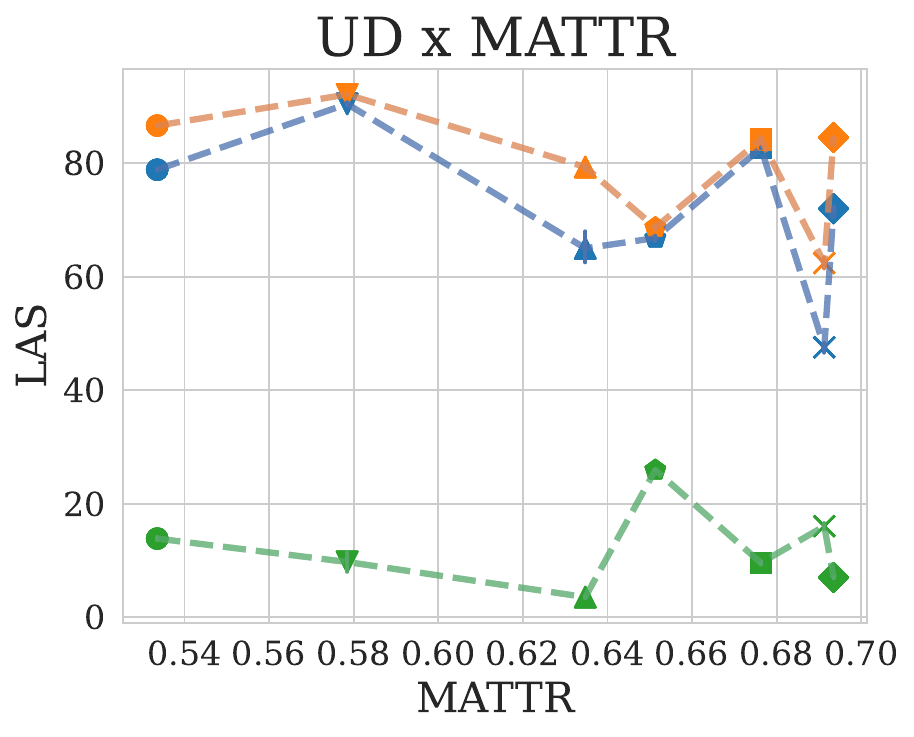}
    \includegraphics[width=0.45\linewidth]{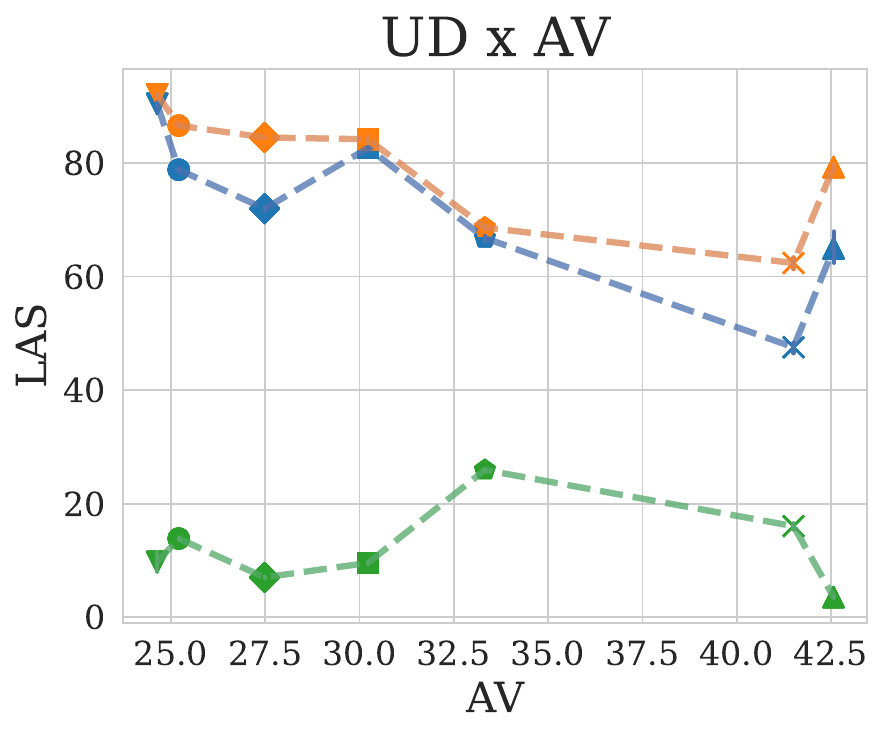}
    
    \includegraphics[width=0.45\linewidth]{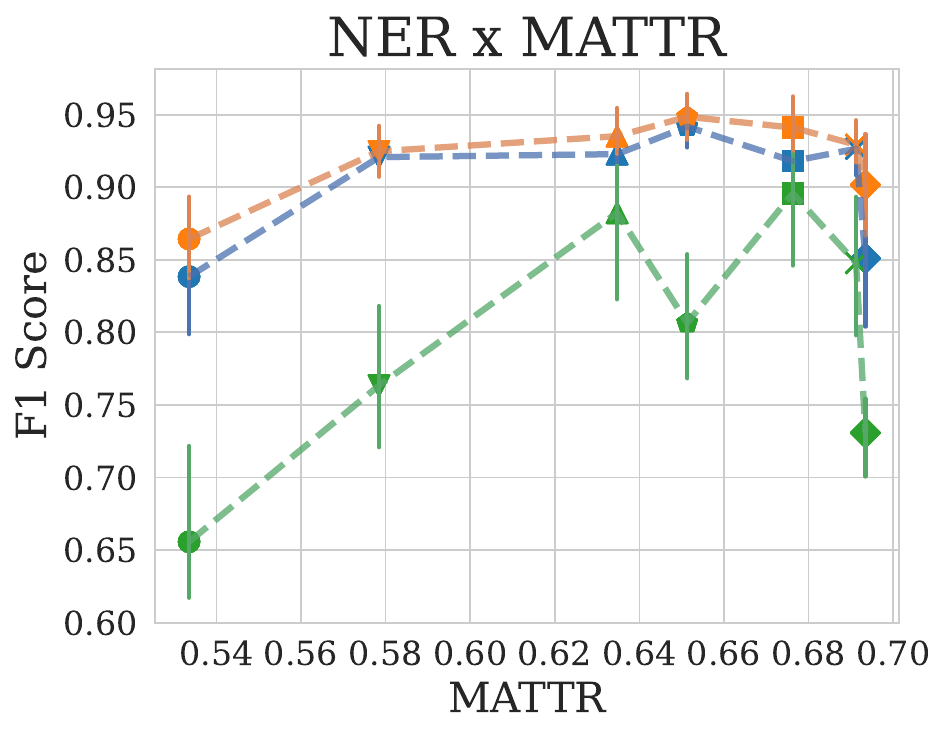}
    \includegraphics[width=0.45\linewidth]{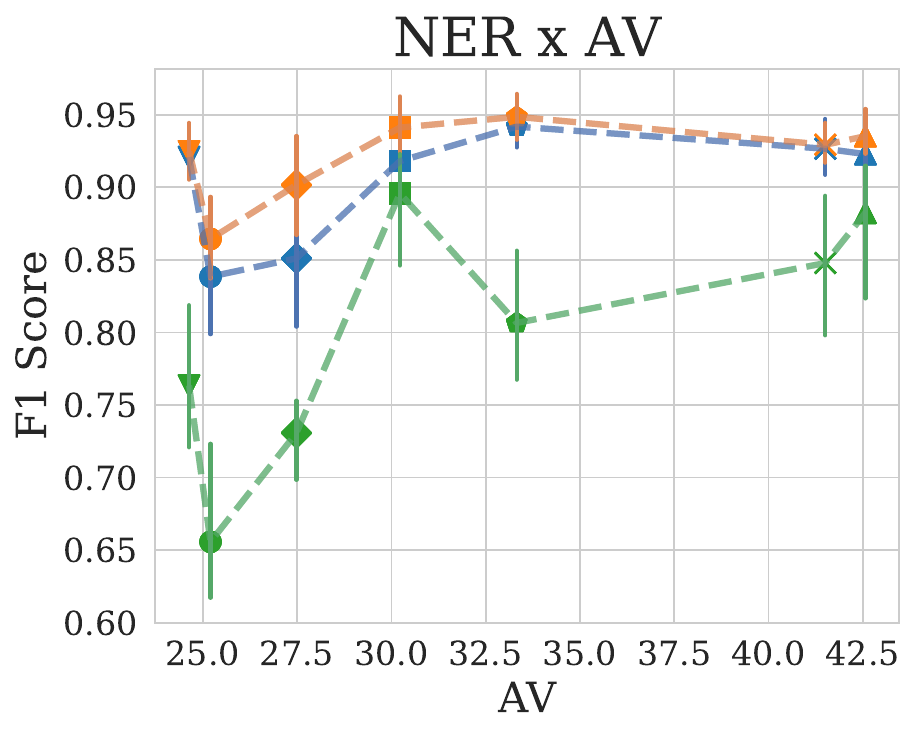}

    \includegraphics[width=0.45\linewidth]{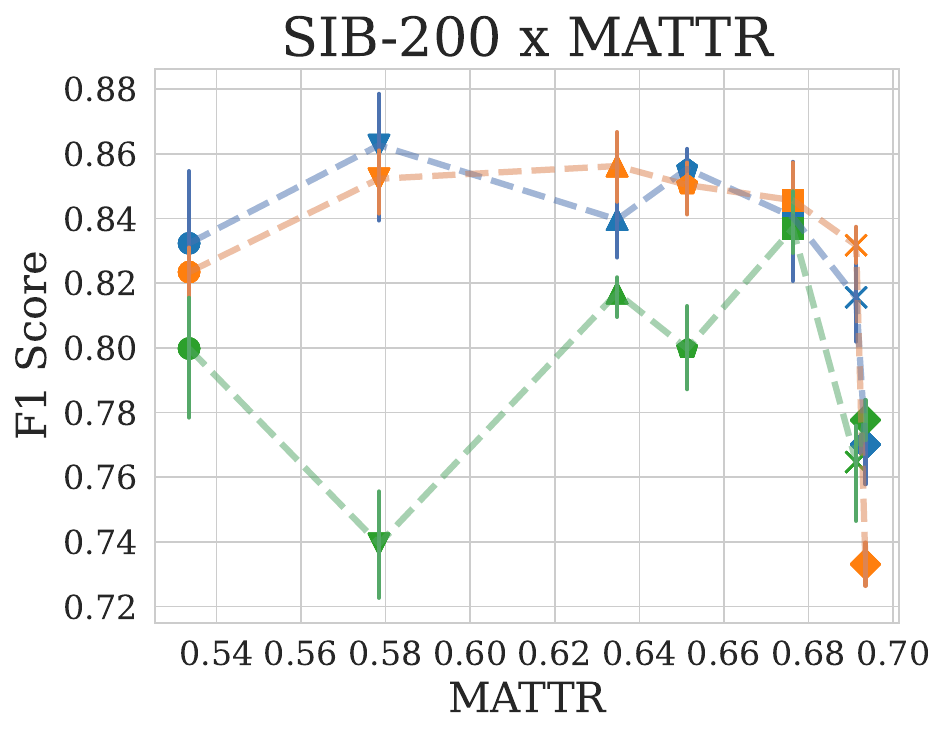}
    \includegraphics[width=0.45\linewidth]{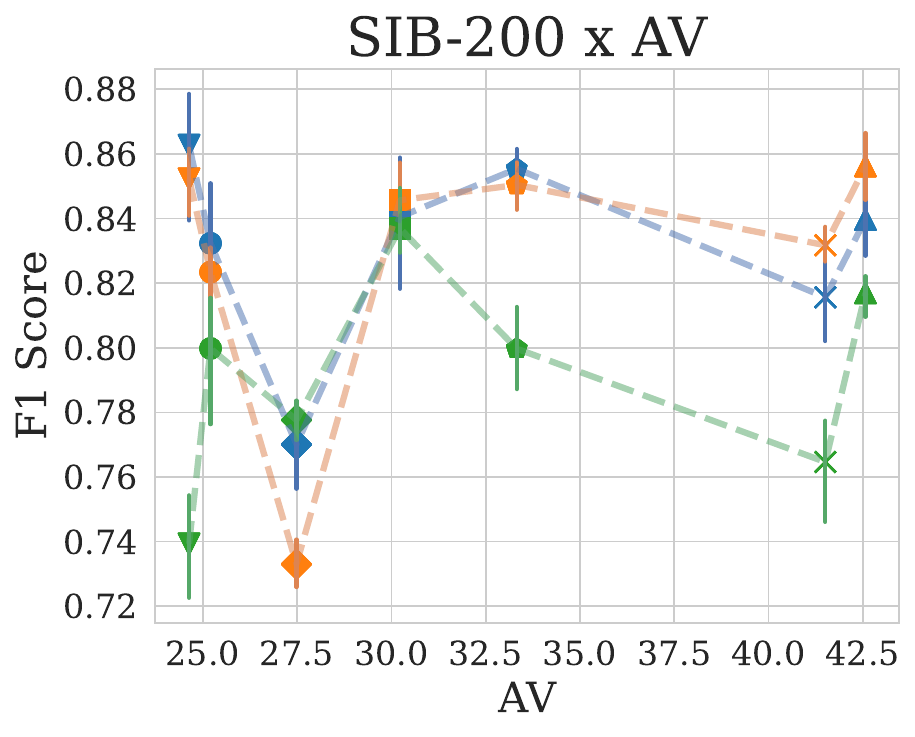}

    \includegraphics[width=0.45\linewidth]{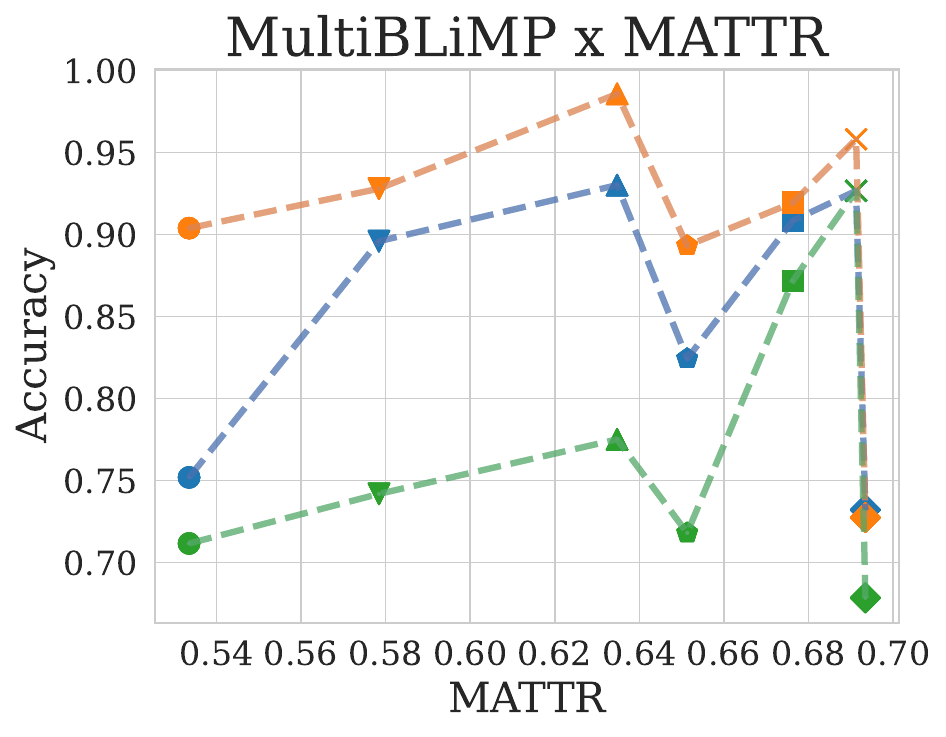}
    \includegraphics[width=0.45\linewidth]{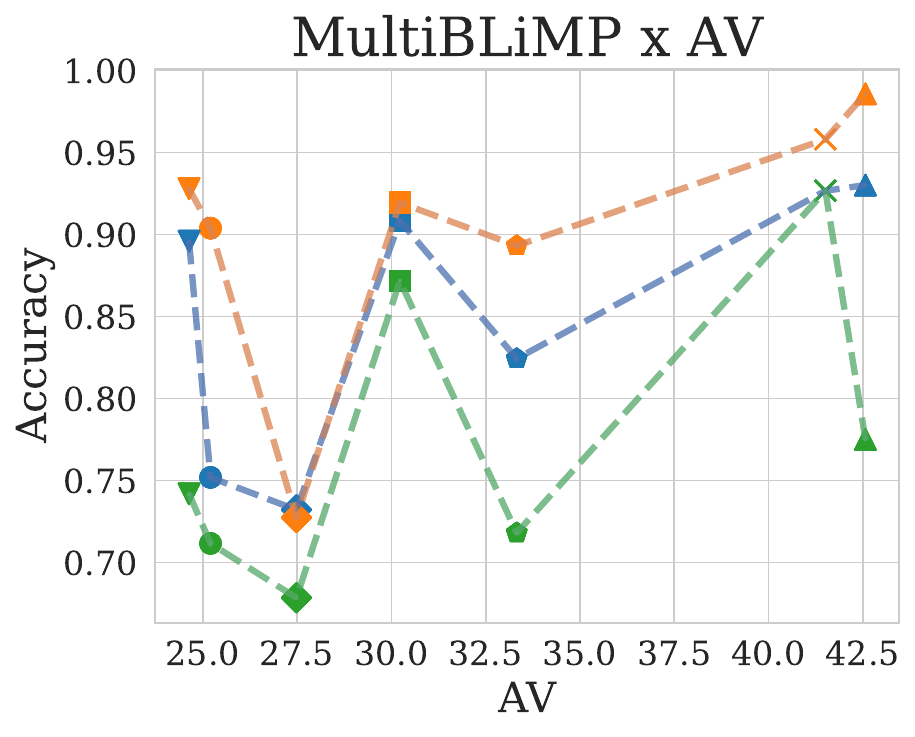}
    
    \includegraphics[width=\linewidth]{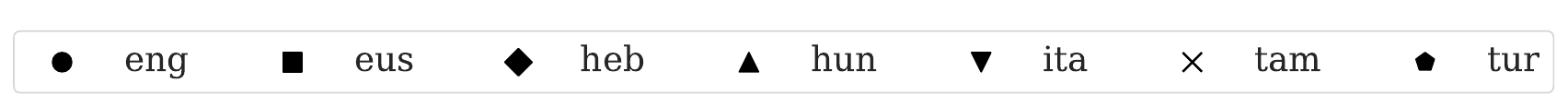}
    \caption{Relation between proxies for morphological complexity and downstream performance. The line shows the groupings of positional encoding type.}
    \label{fig:morph-complexity}
\end{figure}

We attempt to verify the findings of \citet{ghosh2024morphologybased}, who find a significant \emph{negative} correlation between TTR (as a proxy for morphological complexity) and the relative decrease in performance from removing positional encodings per language.
The authors consistently find this across tasks (even with different languages per task).
As mentioned, instead of removing positional encodings of an already trained model, we train tokenizers and models from scratch (Table \ref{tab:comparison}).
Figure \ref{fig:morph-complexity} shows the relation between downstream performance and our proxies for morphological complexity. 

We fail to see the same patterns as \citeauthor{ghosh2024morphologybased}: there is no clear trend from language to language within the tasks. 
A likely explanation for this is that the models used by \citeauthor{ghosh2024morphologybased}, despite having position embedding set to zero post-hoc, still had the benefits of being pre-trained with position encodings. Pretraining from scratch without position encodings, as illustrated in Figure \ref{fig:loss}, results in a model that is not able to converge. 

Moreover, \citeauthor{ghosh2024morphologybased}'s correlations could be attributed to morphologically complex languages potentially being harder to model \cite[\eg][]{mielke2019what,park2021morphology}. This is shown in the relation between UD results and AV and MATTR in Figure \ref{fig:morph-complexity}. The performance with position encodings degrades with an increase in AV and MATTR. The performance without position encodings remains steady, albeit perhaps decreasing slightly with morphological complexity. This results in a smaller gap between \nopos and models trained with position encodings for languages with higher MATTR and AV. \citeauthor{ghosh2024morphologybased} attribute this to the effect of position encodings (this is likely where the significance in their statistical testing comes from), but it is potentially a result of morphologically complex languages performing worse. 

Additionally, these findings are highly dependent on the languages studied and metrics used.
Tamil and Basque, for example have similar MATTR of 0.69, but substantially different AVs: 41.50 and 30.22, respectively.
The lack of a clear trend for \emph{either} metric shows the impact of position encodings on model performance is not dependent upon the morphological complexity of the language.

\begin{figure}[ht]
    \centering
    \includegraphics[width=0.45\linewidth]{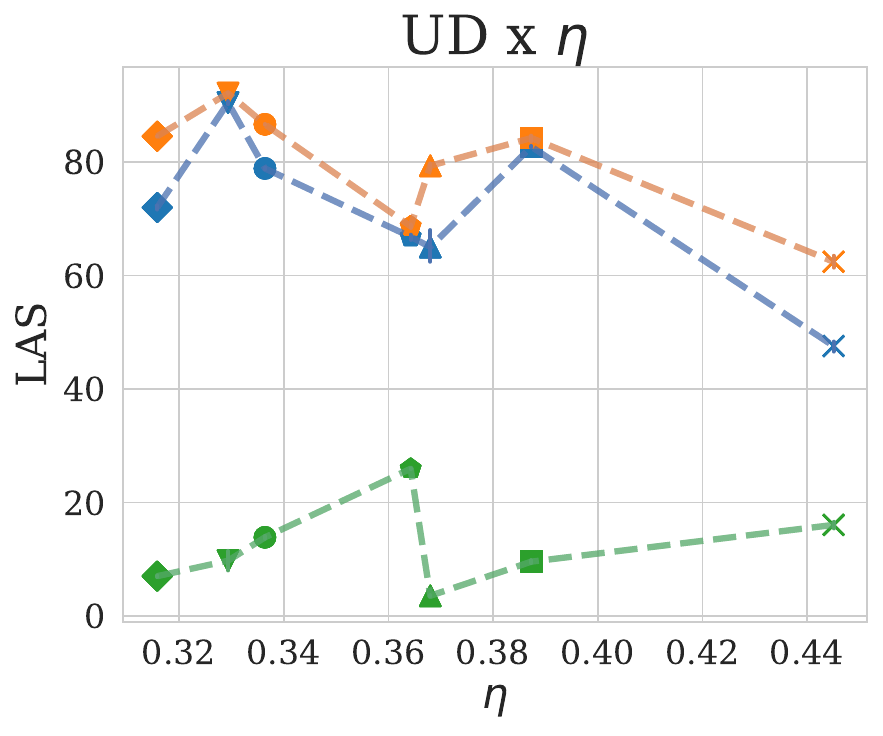}
    \includegraphics[width=0.45\linewidth]{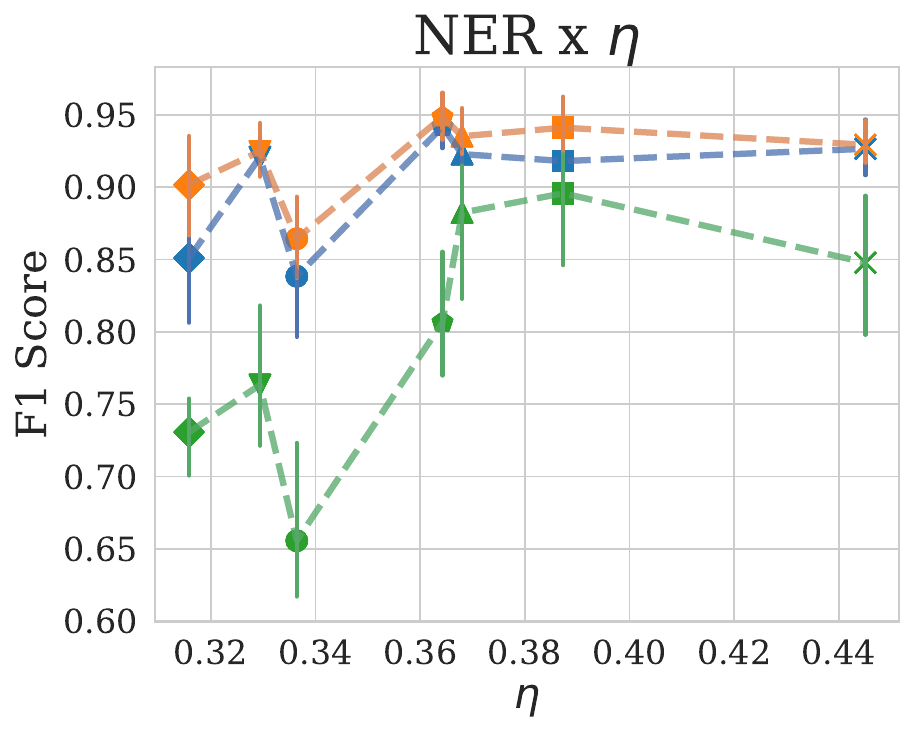}
    \includegraphics[width=0.45\linewidth]{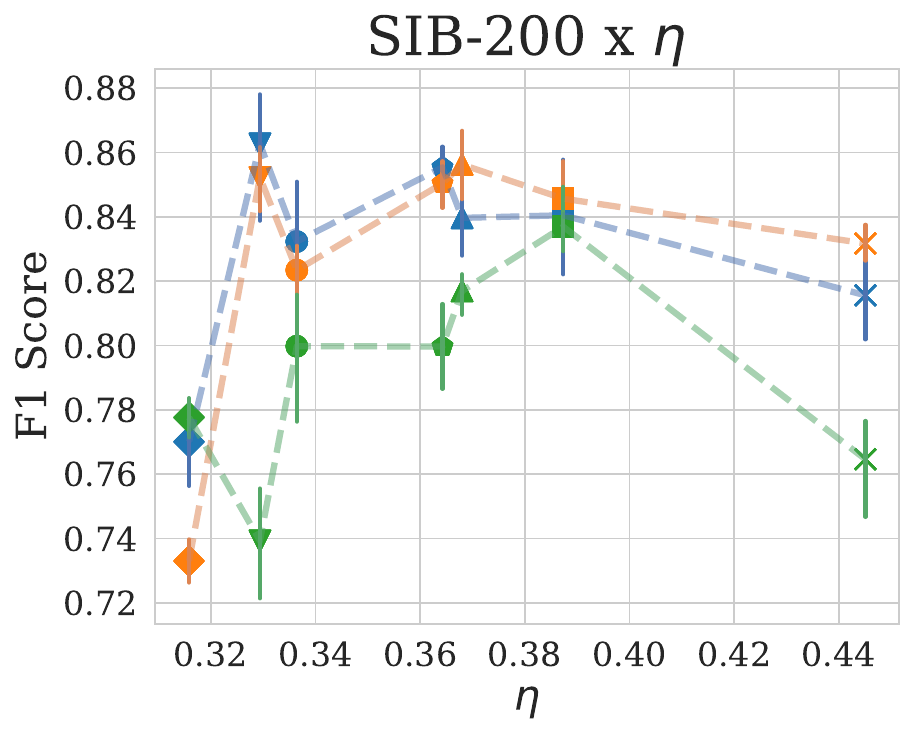}
    \includegraphics[width=0.45\linewidth]{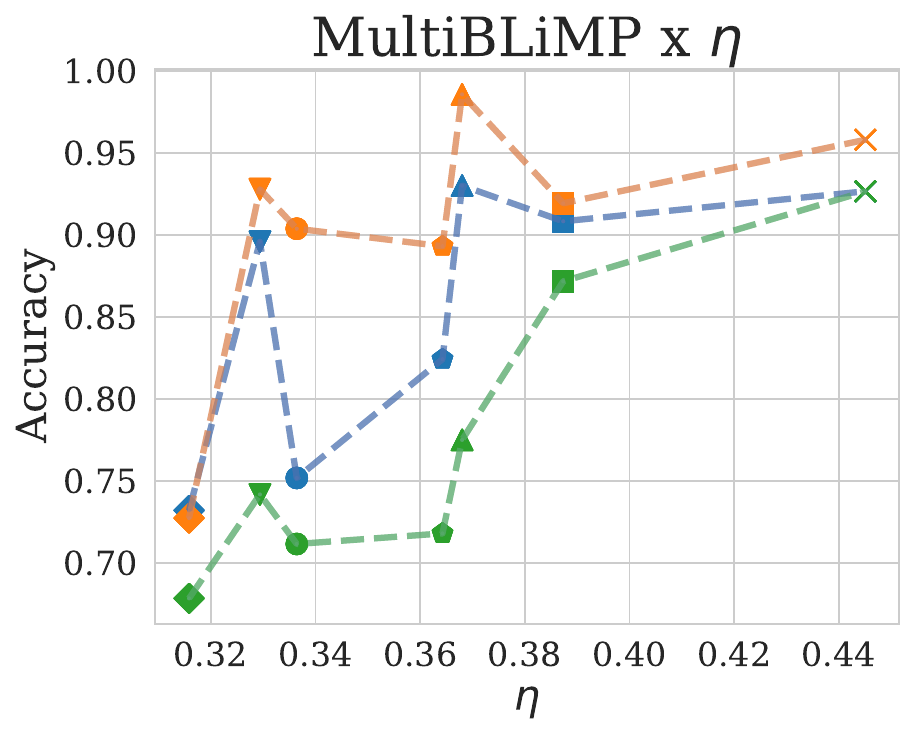}
    \includegraphics[width=\linewidth]{figures/scatterplots/legend.pdf}
    \caption{Relation between entropic efficiency of the accessor variety and downstream performance.}
    \label{fig:efficiency}
\end{figure}

Similarly, we observe comparable AV for both Hungarian and Tamil (42.57 and 41.50, respectively), but the entropic efficiency of their AV distribution ($\eta$) differs substantially: 0.368 and 0.445.
This means that while the number of valid types to choose from is roughly the same from token to token, \emph{how equally valid} those choices are is different (i.e., increased complexity; \S\ref{sec:morph}).
Figure \ref{fig:efficiency} shows the relation between task performance and efficiency, and again we fail to see a clear trend.

\subsection{Position Encodings and Word~Order~Flexibility}
\citet{ghosh2024morphologybased} draw the conclusions that (1) languages with a higher TTR (more morphologically complex) show a smaller performance drop when removing positional encodings compared to those with a lower TTR; that (2) this is caused by morphologically complex languages having a more flexible word order; and therefore (3) languages with a more rigid word order benefit more from positional encodings.

\begin{figure}[ht]
    \centering
    \includegraphics[width=0.45\linewidth]{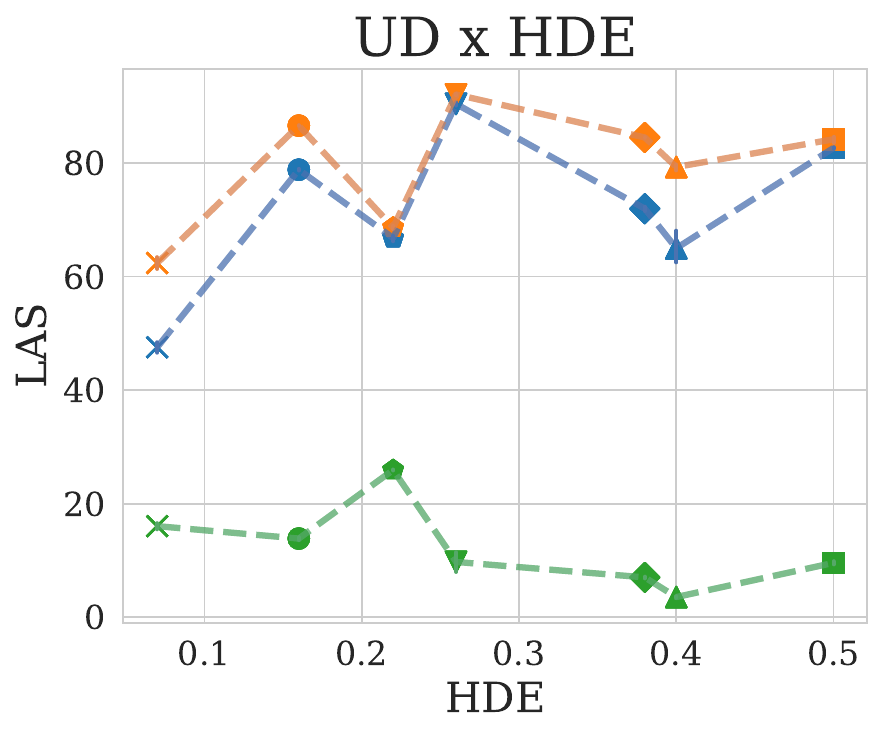}
    \includegraphics[width=0.45\linewidth]{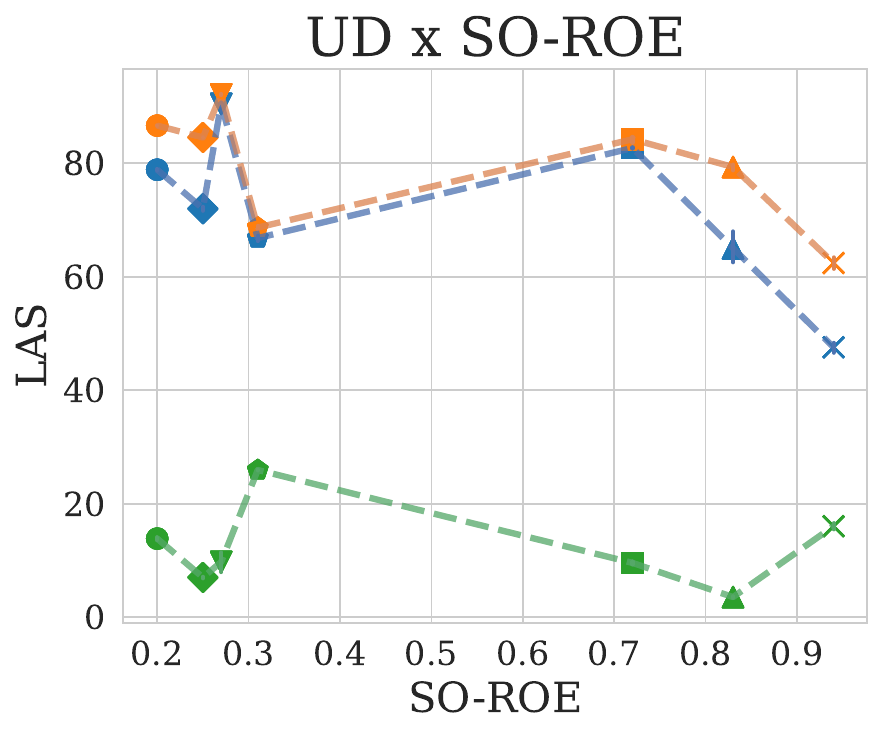}
    
    \includegraphics[width=0.45\linewidth]{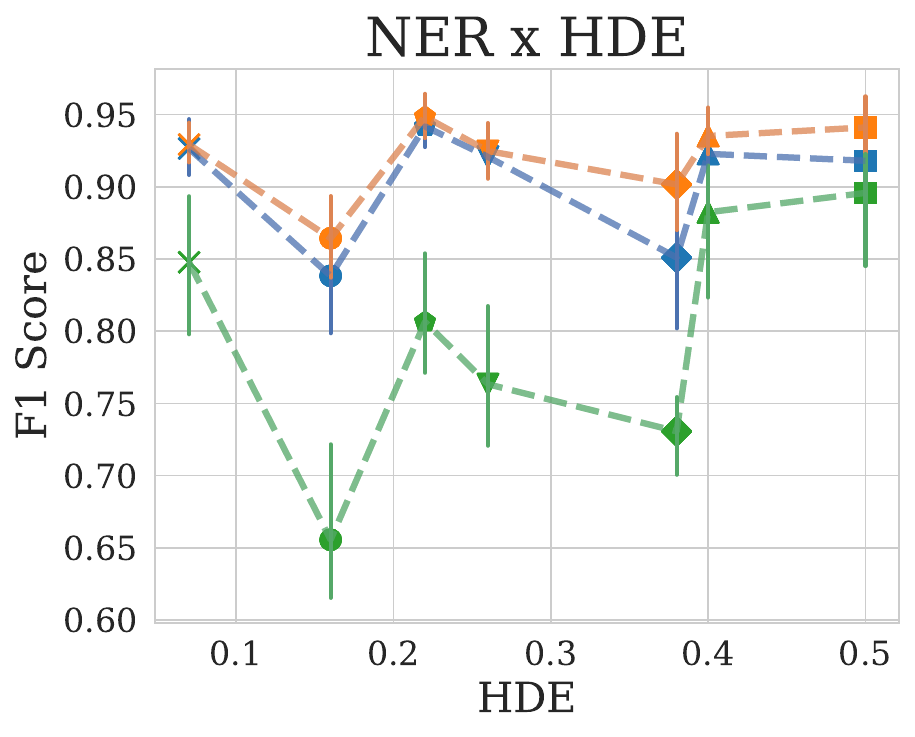}
    \includegraphics[width=0.45\linewidth]{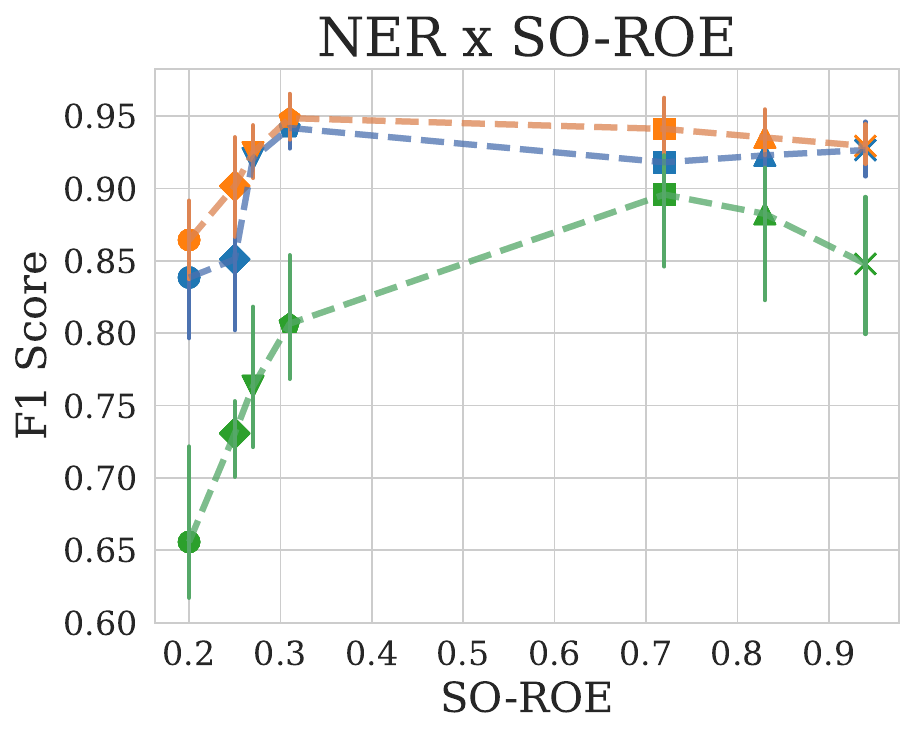}
    
    \includegraphics[width=0.45\linewidth]{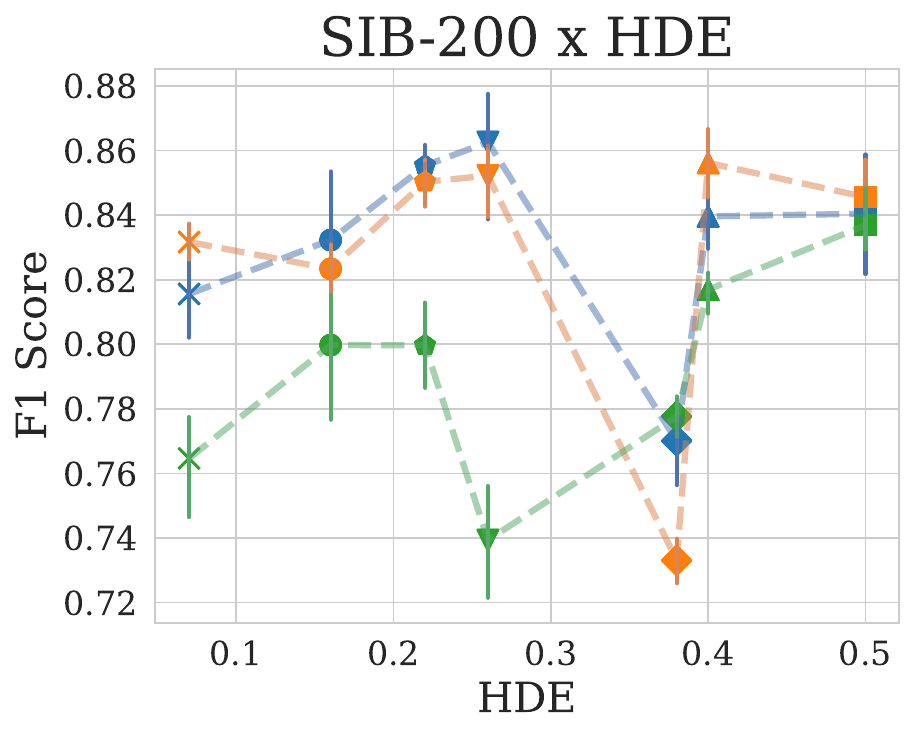}
    \includegraphics[width=0.45\linewidth]{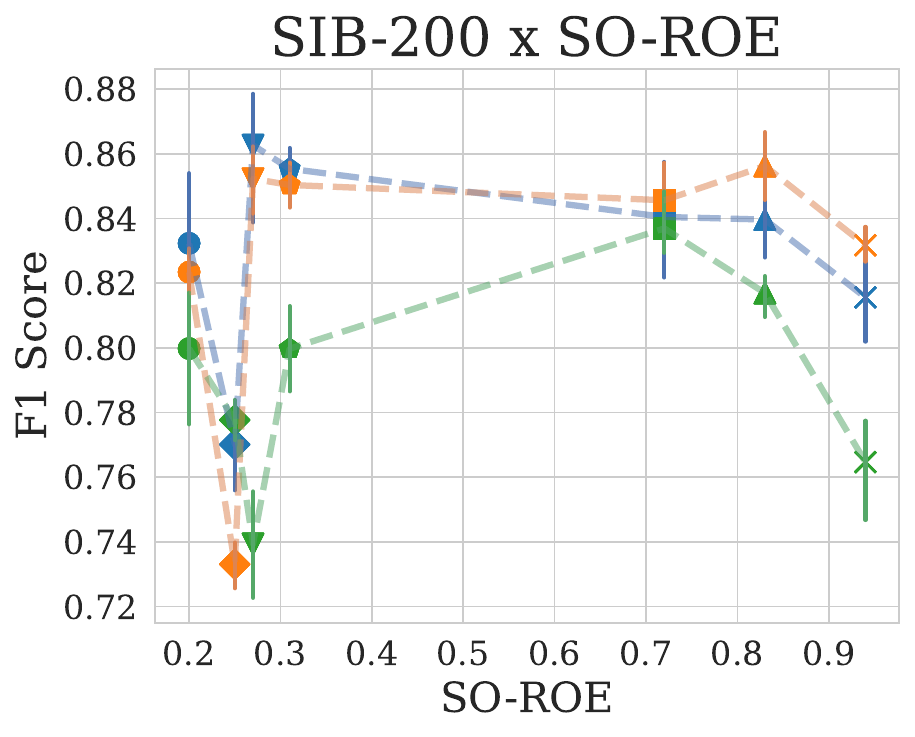}

    \includegraphics[width=0.45\linewidth]{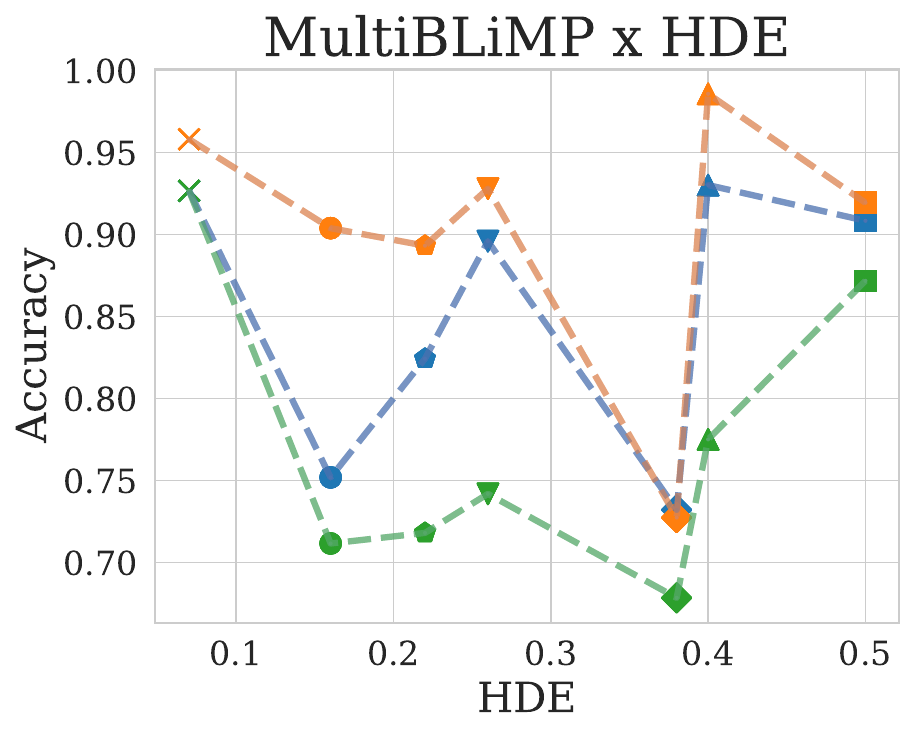}
    \includegraphics[width=0.45\linewidth]{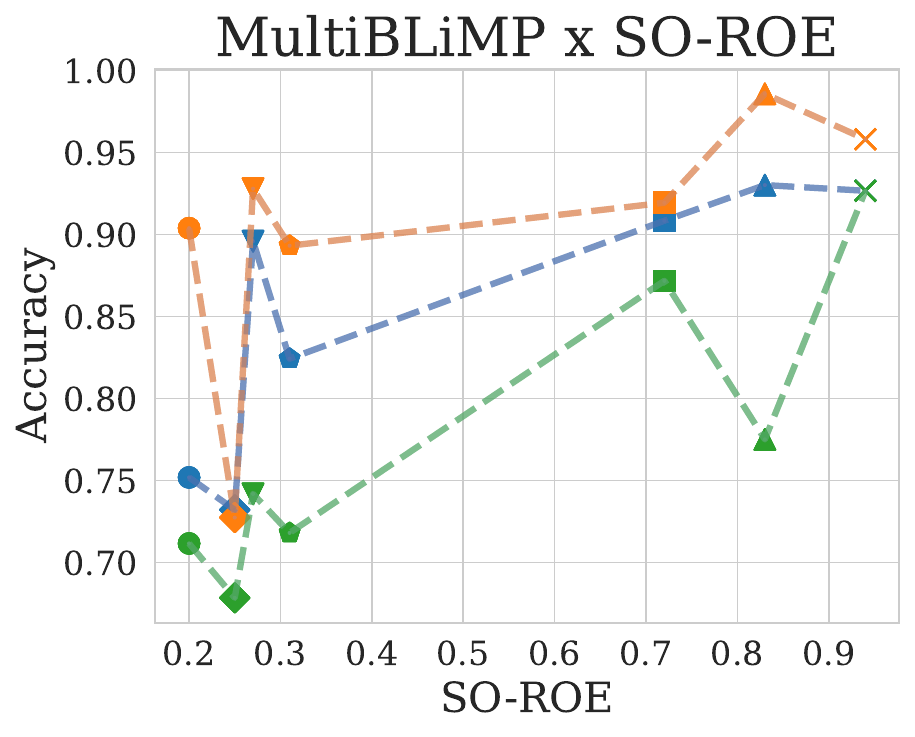}
    
    \includegraphics[width=1\linewidth]{figures/scatterplots/legend.pdf}
    \caption{Relation between proxies for word order flexibility and downstream performance. The connecting line is just to show groupings of positional encoding type, not a regression line.}
    \label{fig:order-flexibility}
\end{figure}

As mentioned in \S\ref{sec:method} and shown in Table \ref{tab:languages}, metrics or proxies for morphological complexity and word order flexibility measure noticeably different phenomena. 
Therefore, drawing conclusions about word order flexibility from TTR alone is problematic. 
To validate their findings, we use gradient word order metrics (\S\ref{sec:method})  and study their relation with downstream performance, shown in Figure \ref{fig:order-flexibility}.

We again fail to see the relation found by \citeauthor{ghosh2024morphologybased}. Our results are strongly dependent on the task, but also on the choice of languages and metrics.
Tamil, for example, has the highest \soroe in our sample, but the \emph{lowest} HDE.
This means that, while it has a very flexible word order for subjects and objects, this is \emph{not} true for general word order flexibility.

As noted in \S\ref{sec:metrics}, there is a mismatch in the granularity of the units used to calculated the metrics (words) and the units seen by a model (subword). Any variations in \emph{word}-order may not be entirely reflected in the \emph{subword}-order of a sentence. For example, a very long word may be able to shift positions in a sentence, but its subwords will have more fixed positions. Thus, perhaps there is a need to adapt these proxies to tokens as well. We leave this investigation to future work. 

\subsection{Position Encodings and Task Performance}
As noted from the results in Figure \ref{fig:results}, the impact of position encodings is task specific. This is also observed by \citeauthor{ghosh2024morphologybased}, who find more variability in performance for syntactic tasks than semantic tasks. Token-level tasks like UD require more position information than sentence-level tasks like \sib. A clear example is seen with Basque. It has a high HDE and \soroe, indicating a language with very flexible word order. One would expect that position information would not matter much in learning its structure. However, with UD we observe a 40 point drop in LAS with \nopos, as compared to \rel, indicating that even with flexible word order, dependency structures are learnt better with position encodings. The variability decreases with NER, since the task does not require much syntactic understanding, but some understanding of token positions in spans is needed. \mblimp has more variability than NER between the position types, even though it is a sentence level task, since understanding of some syntactic structures is essential for predicting the grammaticality of a sentence. \sib has the least variability in scores between \abs, \rel, and \nopos, not being dependent on the learning of syntax. 

Without position encodings, as \citet{wang2020position} note, a transformer model is basically a bag of words model. It does not have enough information to infer syntax, therefore \nopos is still able to compete with \abs and \rel in tasks where the order of tokens is not essential. \citeauthor{wang2020position} found absolute embeddings to outperform relative embeddings for sentence-level classification tasks.
We do not observe this consistently across languages. Notably, with Tamil, Hungarian and Basque, \rel performs better. Thus, in this aspect, conclusions made from English do not carry over to other languages. Perhaps a useful investigation for future work would be to find optimal task-specific position encoding strategies for different languages.  
\citeauthor{wang2020position} also find relative encodings to be better for span prediction tasks, which is reflected in the NER results across languages.

\subsection{Position Encodings and Syntax}
\begin{figure}
    \centering
    \includegraphics[width=\linewidth]{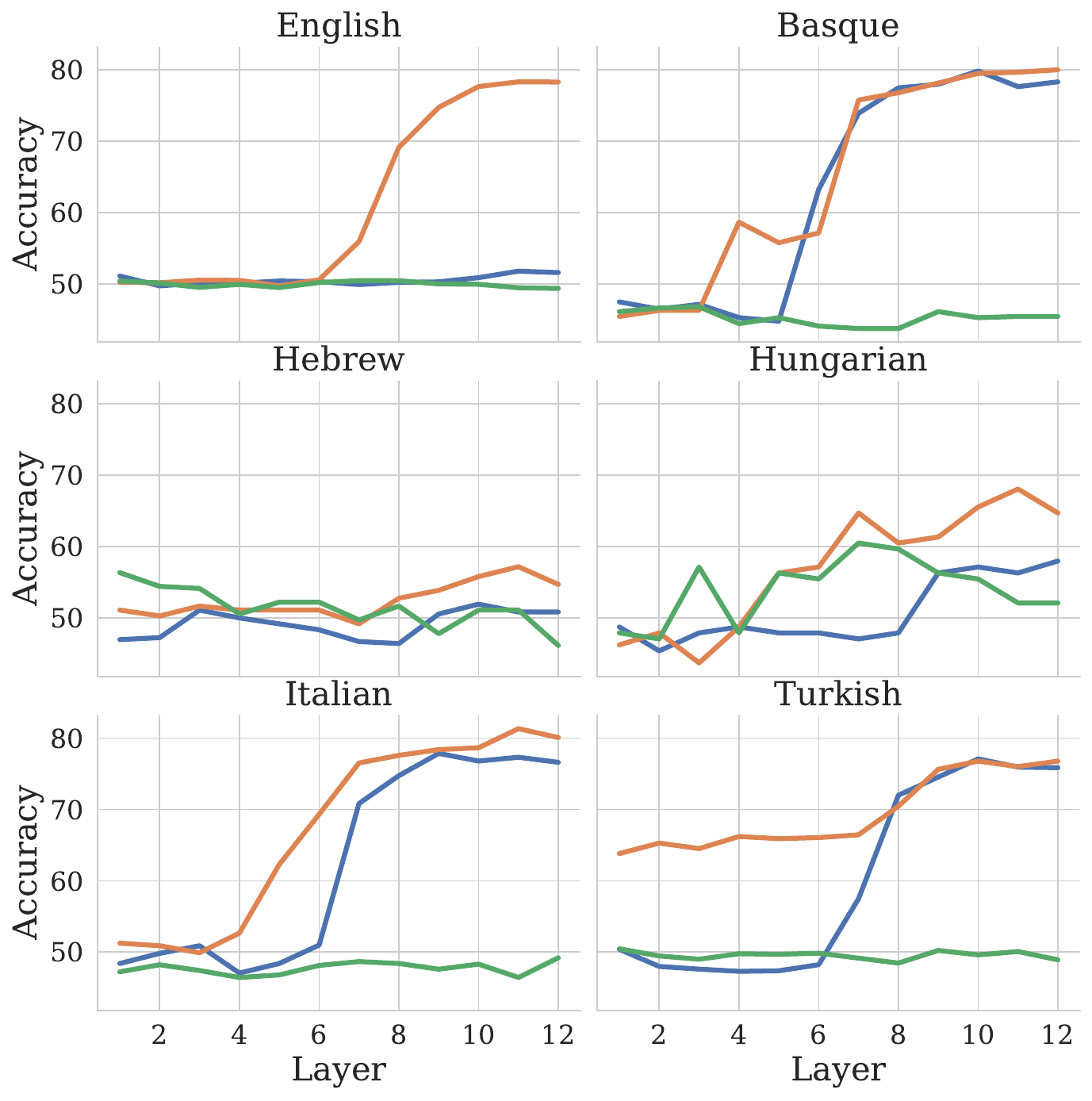}
    \caption{Probing results for \textcolor{abscol}{\abs}, \textcolor{relcol}{\rel}, and \textcolor{nopcol}{\nopos} models. The Tamil dataset generated from UD annotations resulted in only 70 examples, so it is excluded from the plot.}
    \label{fig:probing}
\end{figure}

Results from token-level tasks indicate that position encodings are essential to learning syntax, regardless of typological characteristics. 
We further verify this by probing the models on Bigram Shift, a syntactic task from the \mbox{SentEval} \cite{conneau-etal-2018-cram} framework. We use UD datasets and leverage their annotations to create this probing task for each language \cite{probing-ud}. Using the setup from \citet{tatariya-etal-2024-pixology}, we train logistic regression classifiers on the outputs from each layer of the models to understand whether syntactic information is can be recovered from the hidden representations. Figure \ref{fig:probing} shows the results. The \nopos models do not score higher than the majority baseline of 50\% (Bigram Shift is a binary task) for most languages, clearly indicating that the model does not learn much syntactic information through the layers. Notably, both \abs and \rel for Hebrew, and \abs for English also plateau at 50\%, likely due to the lack of convergence as seen in Figure \ref{fig:loss}. 

\section{Conclusion}\label{sec:conclusion}
We investigate the relation between positional encodings, morphological complexity, and word order flexibility.
We systematically select typologically diverse languages in terms of morphological complexity and word order flexibility.
For each of our seven languages, we pretrain three monolingual models, one with \abs positional encodings, one with \rel, and one without (\nopos).
We use multiple gradient proxies for both morphological complexity and word order flexibility and study their relation with the performance of the different models on various token- and sentence-level downstream tasks.
Our models, tokenizers, and metrics are all based on the same dataset.

We find (1) the impact of position encodings is task specific, (2) relative position encoding gives the most consistent performance across languages and tasks, (3) position encodings are important to learn syntax, regardless of word order flexibility or morphological complexity, and (4) morphological complexity and word order flexibility do not have a direct interaction with position encodings, contradicting previous research.

For future work, we suggest developing a corpus-based proxy for word order flexibility that uses the same units that the model uses (i.e., tokenizer-based), thus removing potential confounds of domain shift and differing units.

\section*{Limitations}\label{sec:limitations}
\paragraph{Language Coverage.}
We systematically select languages based on typological features in order to get a representative sample that covers the language characteristics we are interested in.
Still, it is certainly possible we missed languages simply because they do not have enough data or are not included in the datasets we use.

\paragraph{Metrics.}
While we use gradient proxies for morphological complexity and word order flexibility instead of groupings, they are still \emph{proxies}.
No "perfect" measure for these phenomena exists.

\paragraph{Hyperparameters.}
The \nopos models might be unfairly disadvantaged since we use hyperparameters that were arguably optimized for models with positional encodings.
An exhaustive hyperparameter search was outside the scope and computational budget of this paper.
Our main aim was to keep settings and metrics as comparable as possible.

\paragraph{Models.}
We use encoder-only models to study language phenomena, whereas the field is currently mainly using decoder-only models.
We chose encoder-only models to directly compare findings from \citet{ghosh2024morphologybased}, as well as to leverage the large body of research that investigates positional encodings in these models (see \S\ref{sec:background}).
Future work could look into architecture choices and language phenomena for decoder-only models.

\ifx\review\undefined
\section*{Acknowledgements}
We thank Coleman Haley for pointing us to \citet{futrell2015quantifying}.
We thank the anonymous reviewers for their comments and suggestions.
WP is funded by a KU Leuven Bijzonder Onderzoeksfonds C1 project with reference C14/23/096.
The computational resources and services used were provided by the VSC (Flemish Supercomputer Center), funded by the Research Foundation - Flanders (FWO) and the Flemish Government - department EWI.
\fi

\bibliography{custom}

\clearpage
\appendix

\section{Experimental Setup}\label{app:setup}

\subsection{Language Sampling}\label{app:sampling}
We follow the procedure in \citet{ploeger2025principled} to create a representative language sample of the features we are interested in.
This allows us to make an informed choice about which and how many languages we use given our research question and computational budget.
Our sampling starts with languages that are the intersection of (1) having enough data available in \texttt{FineWeb-2} \cite{penedo2025fineweb2}, (2) being available in our downstream tasks of interest, (3) being available in Grambank \cite{skirgard2023grambank}, and (4) being available in the word order flexibility metrics of \citet{futrell2015quantifying}.
We select the following word order related features from Grambank:

\begin{itemize}
    \item \textsc{GB024}: What is the order of numeral and noun in the NP?
    \item \textsc{GB025}: What is the order of adnominal demonstrative and noun?
    \item \textsc{GB065}: What is the pragmatically unmarked order of adnominal possessor noun and possessed noun?
    \item \textsc{GB130}: What is the pragmatically unmarked order of S and V in intransitive clauses?
    \item \textsc{GB131}: Is a pragmatically unmarked constituent order verb-initial for transitive clauses?
    \item \textsc{GB132}: Is a pragmatically unmarked constituent order verb-medial for transitive clauses?
    \item \textsc{GB133}: Is a pragmatically unmarked constituent order verb-final for transitive clauses?
    \item \textsc{GB134}: Is the order of constituents the same in main and subordinate clauses?
    \item \textsc{GB136}: Is  the order of core argument (i.e., S/A/P) constituents fixed?
    \item \textsc{GB193}: What is the order of adnominal property word and noun?
    \item \textsc{GB203}: What is the order of the adnominal collective universal quantifier ("all") and the noun?
    \item \textsc{GB260}: Can polar interrogation be indicated by a special word order?
\end{itemize}

\begin{figure*}[ht]
    \centering
    \includegraphics[width=0.7\linewidth]{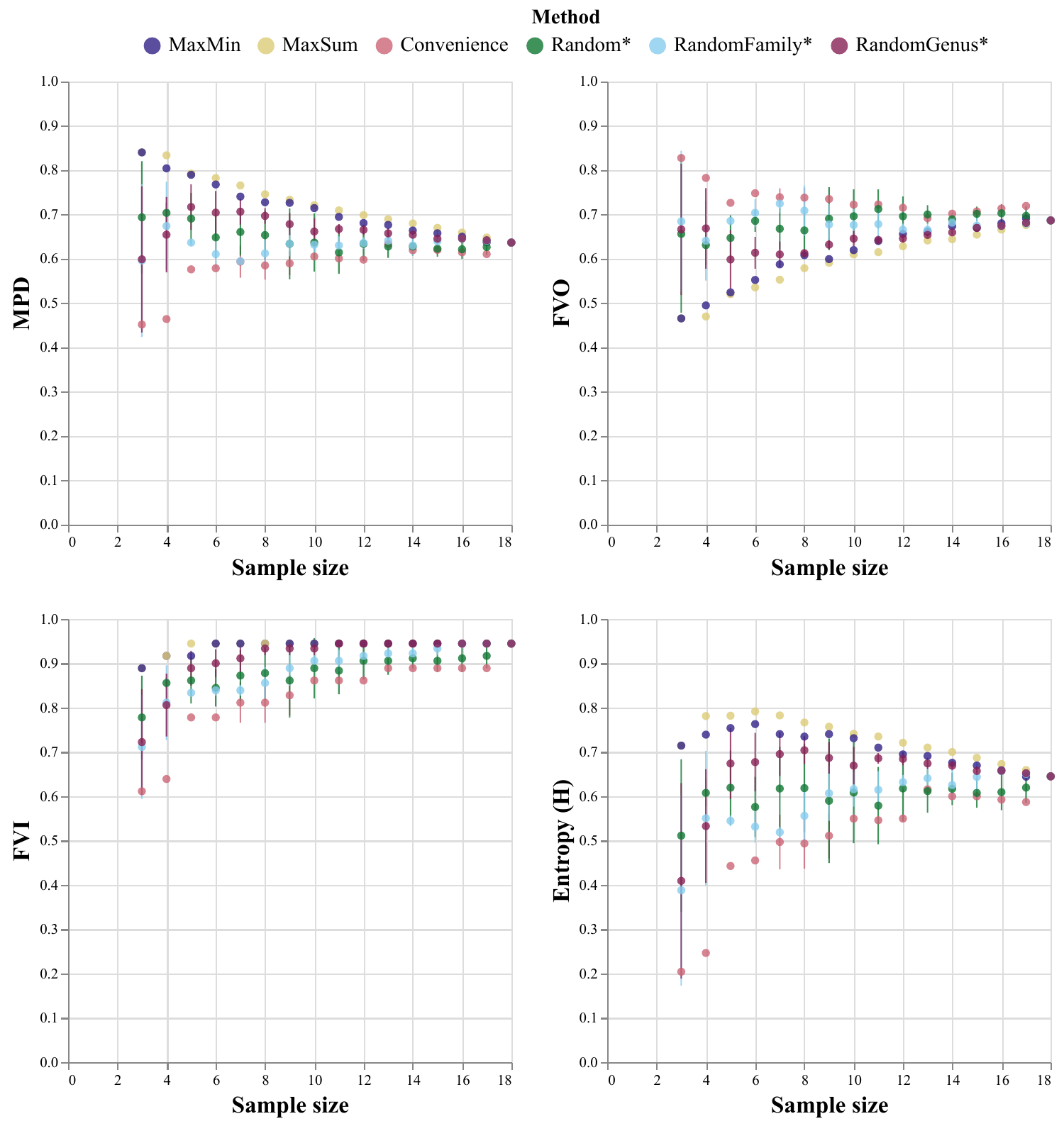}
    \caption{Language sampling metrics for the Grambank features of interest. The features are 
    "saturated" (when looking at FVI and Entropy) with about 6 or 7 languages with the MaxSum objective. $\uparrow$ = higher is better, $\downarrow$ = lower is better. The metrics are: \textbf{MPD} ($\uparrow$) = Mean Pairwise Distance, \textbf{FVI} ($\uparrow$) = Feature Value Inclusion, \textbf{FVO} ($\downarrow$) = Feature Value Overlap, \textbf{Entropy ($H$; $\uparrow$)} = Shannon Entropy of feature vector.}
    \label{fig:lang-sample}
\end{figure*}

These features create a feature vector per language, which is used to calculate pairwise distances.
On these distances, we apply the sampling methods introduced by \citet{ploeger2024what,ploeger2025principled}.
Figure \ref{fig:lang-sample} shows the results of the sampling procedures.
We settle on seven languages that "saturate" (or cover) the features with the MaxSum objective: Basque, English, Hebrew, Hungarian, Italian, Tamil, and Turkish.

\subsection{Model Training}
All pretraining and finetuning hyperparameters are in Tables \ref{tab:pretraining} and \ref{tab:hypers}. We use \texttt{texieve}\footnote{\href{https://github.com/akulmizev/texieve}{github.com/akulmizev/texieve}} for pretraining and finetuning on UD, NER, and \sib. For the tokenizer we used a vocabulary size of 50k.
All the training was run on one NVIDIA H100 GPU. Pretraining took approximately 16 hours for each \abs and \nopos model, and approximately 45 hours for each \rel model, bringing it to a total of 530 hours of training for 21 models. Each finetuning run for UD took approximately 30 minutes (except Tamil, with a small dataset, which took 3 minutes); each \sib run took 6 minutes; each NER run took 25 minutes.
The calculation of the metrics took about 8 hours per language.

\subsection{Datasets}\label{app:datasets}
Table \ref{tab:datasets} lists the details of the train/validation/test split sizes for each finetuning dataset. For WikiAnn we used the version of the dataset available on HuggingFace \cite{wolf2020transformers} under \texttt{unimelb-nlp/wikiann}. For languages having more than one UD datasets, we select the ones with the highest quality as reported in \citet{kulmizev-nivre-2023-investigating}.

\clearpage

\begin{table}[ht]
    \centering
    \resizebox{\linewidth}{!}{\begin{tabular}{lrr}\toprule
\multicolumn{2}{c}{\textbf{Pretraining}} \\\midrule
Model Type &roberta \\
Batch Size &128 \\
Max Length &512 \\
Train Steps &150k \\
Eval Steps &5k \\
Learning Rate & $2\times10^{-4}$ \\
Padding Strategy &longest \\
Mask Probability &0.12 \\
Weight Decay &0.05 \\
Gradient Accumulation Steps &4 \\
Warmup Steps &1500 \\
\bottomrule
\end{tabular}
}
    \caption{Pretraining hyperparameters.}
    \label{tab:pretraining}
\end{table}

\begin{table}[ht]
    \centering
    \resizebox{0.8\linewidth}{!}{\begin{tabular}{lrr}\toprule
\multicolumn{2}{c}{\textbf{Finetuning}} \\\midrule
\multicolumn{2}{c}{UD} \\\midrule
Max Length &512 \\
Epochs &20 \\
Batch Size &16 \\
Learning Rate &$5\times10^{-5}$\\
Padding Strategy &max\_length \\
Warmup Steps &500 \\\midrule
\multicolumn{2}{c}{NER} \\\midrule
Max Length &512 \\
Epochs &5 \\
Batch Size &32 \\
Learning Rate &$5\times10^{-5}$  \\
Padding Strategy &max\_length \\
Warmup Steps &500 \\\midrule
\multicolumn{2}{c}{\sib} \\\midrule
Max Length &512 \\
Epochs &20 \\
Batch Size &32 \\
Learning Rate &$5\times10^{-5}$ \\
Padding Strategy &max\_length \\
Warmup Steps &500 \\
\bottomrule
\end{tabular}
}
    \caption{Finetuning hyperparameters.}
    \label{tab:hypers}
\end{table}

\begin{table}[ht]
    \centering
    \resizebox{\linewidth}{!}{\begin{tabular}{lrrrr}\toprule
\multicolumn{4}{c}{\textbf{Dataset Splits}} \\\cmidrule{1-4}
&\textbf{Train} &\textbf{Validation} &\textbf{Test} \\\midrule
UD\_Basque-BDT &5396 &1798 &1799 \\
UD\_English-EWT &12544 &2001 &2077 \\
UD\_Hebrew-HTB &5168 &484 &491 \\
UD\_Hungarian-Szeged &910 &441 &449 \\
UD\_Italian-IST &13121 &564 &482 \\
UD\_Tamil-TTB &400 &80 &120 \\
UD\_Turkish-Kenet &15398 &1646 &1643 \\\midrule
wikiann-eu &10k &10k &10k \\
wikiann-en &20k &10k &10k \\
wikiann-he &20k &10k &10k \\
wikiann-hu &20k &10k &10k \\
wikiann-it &20k &10k &10k \\
wikiann-ta &15k &1k &1k \\
wikiann-tr &20k &10k &10k \\\midrule
SIB-200 &701 &99 &99 \\\midrule
MultiBLiMP-eus &- &- &278 \\
MultiBLiMP-eng &- &- &770 \\
MultiBLiMP-heb &- &- &2.3k \\
MultiBLiMP-hun &- &- &843 \\
MultiBLiMP-ita &- &- &3k \\
MultiBLiMP-tam &- &- &382 \\
MultiBLiMP-tur &- &- &1.7k \\

\bottomrule
\end{tabular}
}
    \caption{Dataset statistics. \mblimp has a single split named "train" on Huggingface, but it is used as a test split.}
    \label{tab:datasets}
\end{table}

\clearpage
\section{Additional Results}\label{app:additional-results}
\subsection{UD and Fertility}
\begin{figure}[ht]
    \centering
    \includegraphics[width=1\linewidth]{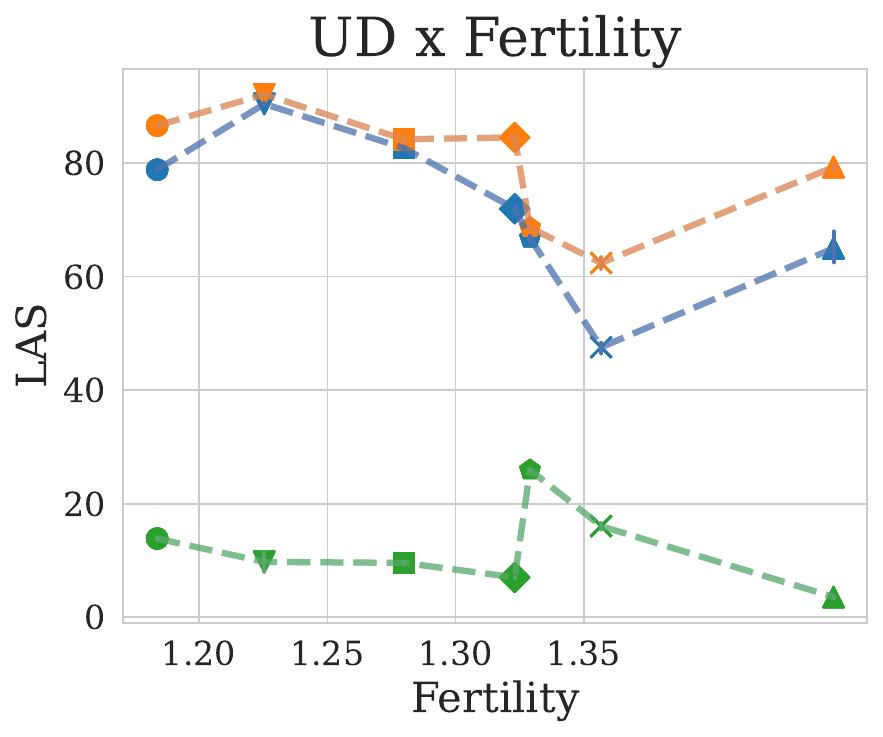}
    \caption{Relation between tokenizer fertility and LAS for UD. The connecting line shows the groupings of positional encoding type. Fertility is calculated on the test split of \texttt{FineWeb-2} (and a similar-sized sample of \texttt{FineWeb} for English that was also used for validation of the models).}
    \label{fig:fertility}
\end{figure}

To better understand UD results, we look at \citet{rust-etal-2021-good} who found a strong negative correlation between tokenizer fertility and performance on UD. We calculate fertility for all our tokenizers and plot their relation with LAS in Figure \ref{fig:fertility}.
We can observe a slight downward trend for \abs and \rel scores with increase in fertility. This trend is notably absent with \nopos models.

Future work can look at the relation between positional encodings and tokenization.
There are many interesting areas to explore in that direction, but they are outside of the scope of our current study.
\clearpage
\onecolumn
\subsection{Full Results}\label{app:full-results}

\begin{table*}[ht]
    \centering
    \begin{tabular}{lrrrrc}\toprule
\textbf{Language} &\textbf{Pos Type} & \multicolumn{1}{c}{\textbf{UD}} &\multicolumn{1}{c}{\textbf{WikiAnn}} &\multicolumn{1}{c}{\textbf{SIB-200}} &\multicolumn{1}{c}{\textbf{MultiBLiMP}} \\\midrule
English &\nopos &13.87$_{\pm0.32}$ &0.66$_{\pm0.07}$ &0.80$_{\pm0.02}$ &0.71 \\
&\abs &78.84$_{\pm0.18}$ &0.84$_{\pm0.05}$ &\textbf{0.83$_{\pm0.02}$} &0.75 \\
&\rel &\textbf{86.61$_{\pm0.07}$} &\textbf{0.86$_{\pm0.03}$} &0.82$_{\pm0.01}$ &\textbf{0.90} \\\midrule
Basque &\nopos &9.59$_{\pm0.33}$ &0.90$_{\pm0.05}$ &0.84$_{\pm0.01}$ &0.87 \\
&\abs &82.67$_{\pm0.15}$ &0.92$_{\pm0.01}$ &0.84$_{\pm0.02}$ &0.91 \\
&\rel &\textbf{84.20$_{\pm0.71}$} &\textbf{0.94$_{\pm0.02}$} &\textbf{0.85$_{\pm0.01}$} &\textbf{0.92} \\\midrule
Hebrew &\nopos &7.03$_{\pm0.18}$ &0.73$_{\pm0.03}$ &\textbf{0.78$_{\pm0.01}$} &0.68 \\
&\abs &71.97$_{\pm0.48}$ &0.85$_{\pm0.06}$ &0.77$_{\pm0.01}$ &\textbf{0.73} \\
&\rel &\textbf{84.52$_{\pm0.06}$} &\textbf{0.90$_{\pm0.04}$} &0.73$_{\pm0.01}$ &\textbf{0.73} \\\midrule
Hungarian &\nopos &3.57$_{\pm0.18}$ &0.73$_{\pm0.03}$ &0.82$_{\pm0.01}$ &0.78 \\
&\abs &65.00$_{\pm3.34}$ &0.85$_{\pm0.06}$ &0.84$_{\pm0.01}$ &0.93 \\
&\rel &\textbf{79.31$_{\pm0.67}$} &\textbf{0.90$_{\pm0.04}$} &\textbf{0.86$_{\pm0.01}$} &\textbf{0.99} \\\midrule
Italian &\nopos &9.75$_{\pm2.27}$ &0.76$_{\pm0.06}$ &0.74$_{\pm0.02}$ &0.74 \\
&\abs &90.42$_{\pm0.36}$ &\textbf{0.92$_{\pm0.02}$} &\textbf{0.86$_{\pm0.02}$} &0.90 \\
&\rel &\textbf{92.09$_{\pm0.15}$} &\textbf{0.92$_{\pm0.02}$} &0.85$_{\pm0.01}$ &\textbf{0.93} \\\midrule
Tamil &\nopos &16.05$_{\pm0.71}$ &0.85$_{\pm0.06}$ &0.76$_{\pm0.02}$ &0.93 \\
&\abs &47.57$_{\pm1.16}$ &\textbf{0.93$_{\pm0.02}$} &0.82$_{\pm0.02}$ &0.93 \\
&\rel &\textbf{62.40$_{\pm1.32}$} &\textbf{0.93$_{\pm0.02}$} &\textbf{0.83$_{\pm0.01}$} &\textbf{0.96} \\\midrule
Turkish &\nopos &25.95$_{\pm0.10}$ &0.81$_{\pm0.05}$ &0.80$_{\pm0.02}$ &0.72 \\
&\abs &66.80$_{\pm0.60}$ &0.94$_{\pm0.02}$ &\textbf{0.86$_{\pm0.01}$} &0.82 \\
&\rel &\textbf{68.65$_{\pm0.40}$} &\textbf{0.95$_{\pm0.02}$} &0.85$_{\pm0.01}$ &\textbf{0.89} \\
\bottomrule
\end{tabular}

    \caption{Results for UD (LAS), WikiAnn NER ($F_1$),\sib ($F_1$), and \mblimp (accuracy). UD, NER, and \sib scores are an average of 5 runs with different random seeds. Results are the same as in Figure \ref{fig:results}.}
    \label{tab:results}
\end{table*}

\end{document}